\documentclass[10pt,twocolumn]{article}

\usepackage[letterpaper, margin=0.75in, columnsep=0.25in]{geometry}

\usepackage[utf8]{inputenc}
\usepackage[T1]{fontenc}
\usepackage{textcomp}   

\usepackage{times}
\usepackage{microtype}

\usepackage{titlesec}
\titleformat{\section}{\normalfont\large\bfseries}{\thesection}{0.5em}{}
\titleformat{\subsection}{\normalfont\normalsize\bfseries}{\thesubsection}{0.5em}{}
\titleformat{\subsubsection}{\normalfont\small\bfseries}{\thesubsubsection}{0.5em}{}
\titlespacing{\section}{0pt}{1.0em}{0.4em}
\titlespacing{\subsection}{0pt}{0.7em}{0.3em}
\titlespacing{\subsubsection}{0pt}{0.5em}{0.2em}

\usepackage{booktabs}
\usepackage{graphicx}
\usepackage{caption}
\usepackage{enumitem}
\usepackage{multirow}
\usepackage{tabularx}
\usepackage{xcolor}
\usepackage{url}
\usepackage{amsmath}
\usepackage{float}
\usepackage[round]{natbib}
\usepackage{placeins}  

\usepackage{hyperref}
\hypersetup{
    colorlinks=true,
    linkcolor={blue!60!black},
    citecolor={blue!60!black},
    urlcolor={blue!80!black},
    pdfauthor={Jaden Zhang and Gardenia Liu and Oliver Johansson and Hileamlak Yitayew and Kamryn Ohly and Grace Li},
    pdftitle={Prediction Arena: Benchmarking AI Models on Real-World Prediction Markets},
    pdfkeywords={AI benchmarks, prediction markets, language models, autonomous agents, real-world evaluation},
}

\newcommand{\benchmark}{Prediction Arena}
\newcommand{\kalshi}{Kalshi}
\newcommand{\polymarket}{Polymarket}
\newcommand{\model}[1]{\texttt{#1}}
\title{\textbf{Prediction Arena:}\\
       \large Benchmarking AI Models on Real-World Prediction Markets}

\author{
  \small
  Jaden Zhang$^{1,2}$ \quad
  Gardenia Liu$^{1,2}$ \quad
  Oliver Johansson$^{1}$ \quad
  Hileamlak Yitayew$^{1}$ \quad
  Kamryn Ohly$^{1}$ \quad
  Grace Li$^{1}$ \\[0.3em]
  \footnotesize $^1$Arcada Labs \quad $^2$Harvard University \\[0.15em]
  \footnotesize\textit{Evaluation period: January 12 -- March 9, 2026}
}

\date{March 28, 2026}

\begin{document}

\maketitle

\begin{abstract}
  We introduce \benchmark{}, a benchmark for evaluating AI models' predictive accuracy and decision-making by enabling them to trade autonomously on real prediction markets with real capital. Unlike synthetic benchmarks, \benchmark{} tests models in live environments where trades execute on actual exchanges (\kalshi{} and \polymarket{}), providing objective ground truth that cannot be gamed or overfitted. Each model operates as an independent agent starting with \$10,000, making autonomous decisions every 15--45 minutes.

Over a 57-day longitudinal evaluation (January 12 to March 9, 2026), we track two cohorts: six frontier models in live trading (\textbf{Cohort 1}, full period) and four next-generation models in paper trading (\textbf{Cohort 2}, 3-day preliminary). For the primary Cohort 1, final \kalshi{} returns range from $-16.0\%$ to $-30.8\%$. Our analysis identifies a clear performance hierarchy: initial prediction accuracy and the ability to capitalize on correct predictions are the main drivers, while research volume shows no correlation with outcomes.

A striking cross-platform contrast emerges from parallel \polymarket{} live trading: Cohort 1 models averaged only $-1.1\%$ on \polymarket{} vs.\ $-22.6\%$ on \kalshi{}, with \model{grok-4-20-checkpoint} achieving a $71.4\%$ settlement win rate — the highest across any platform or cohort. \model{gemini-3.1-pro-preview} (Cohort 2), which executed zero trades on \kalshi{}, achieved $+6.02\%$ on \polymarket{} in 3 days — the best return of any model across either cohort — demonstrating that platform design has a profound effect on which models succeed.

Beyond performance, we analyze computational efficiency (token usage, cycle time), settlement accuracy, exit patterns, and market preferences, providing a comprehensive view of how frontier models behave under real financial pressure.

\end{abstract}

\section{Introduction}

As AI models become increasingly capable of autonomous decision-making, there is a critical need for benchmarks that evaluate their performance in real-world environments with real consequences. Existing benchmarks often rely on synthetic tasks or simulated environments, creating a sim-to-real gap that limits their ability to measure true model capabilities~\citep{swebench, terminalbench, apexagents}. Prediction markets offer a unique opportunity to bridge this gap: they have a long-established track record as accurate information aggregation mechanisms~\citep{wolfers2004prediction, berg2008prediction}, provide objective outcome ground truth, and expose models to real financial consequences. Recent work has shown that LLM ensembles can approach human crowd forecasting accuracy~\citep{schoenegger2024wisdom}, and early forecasting benchmarks have begun probing this capability~\citep{zou2022autocast} — yet no prior work has evaluated frontier models as fully autonomous agents trading with real capital in live prediction markets.

\benchmark{} addresses several limitations of existing benchmarks. First, it evaluates Cohort 1 models with real capital on prediction market exchanges (\kalshi{} and \polymarket{}), ensuring that performance reflects genuine predictive ability rather than overfitting to test sets. Second, it measures models' ability to reason about future events across diverse domains—financial markets, economic indicators, weather, sports, politics, and more—testing general intelligence rather than narrow expertise. Third, it provides objective, quantitative metrics (account value, PnL, win rate) that cannot be gamed through prompt engineering or dataset contamination.

We present two platform implementations that serve different evaluation goals. The \kalshi{} version uses a curated set of 29 standardized markets, ensuring all models see the same opportunities and focusing evaluation on prediction accuracy rather than market selection. The \polymarket{} version provides full market discovery capabilities, allowing agents to search the entire market universe and identify profitable opportunities — a capability that may be crucial for real-world deployment.

Our evaluation spans 57 days (January 12 to March 9, 2026) and includes two cohorts of models. Cohort 1 (six frontier models, live trading, full period) produced final returns ranging from $-16.0\%$ to $-30.8\%$ on \kalshi{}, with a \$1,473 spread between the best and worst performers. Starting February 9, all Cohort 1 models ran live trading on \polymarket{}, where losses were smaller (average $-1.1\%$ vs.\ $-22.6\%$ on \kalshi{}), and \model{grok-4-20-checkpoint} achieved a $71.4\%$ settlement win rate — the highest across any platform or cohort. Cohort 2 (four next-generation models, paper trading, Mar 6–9) reveals that \model{gemini-3.1-pro-preview} — which executed zero trades on \kalshi{} — achieved $+6.02\%$ on \polymarket{} in just 3 days.

Beyond performance metrics, \benchmark{} enables analysis of efficiency measures that are often overlooked in benchmark evaluations. We examine token usage patterns, processing time per decision, research intensity, exit patterns, and trading behavior — providing a comprehensive view of model capabilities that goes beyond simple win rates. This analysis reveals that computational effort does not predict performance: the most capital-efficient model was not the most computationally intensive.

The remainder of this paper is organized as follows: \S\ref{sec:related} reviews related work; \S\ref{sec:benchmark} defines the benchmark design and evaluation philosophy; \S\ref{sec:system} describes the system architecture; \S\ref{sec:methodology} and \S\ref{sec:experiments} detail the methodology and experimental setup; \S\ref{sec:results} reports results across both cohorts and platforms; \S\ref{sec:analysis} provides in-depth behavioral analysis; and \S\ref{sec:limitations}--\ref{sec:conclusion} cover limitations, discussion, and conclusions.

\section{Related Work}
\label{sec:related}

\textbf{Prediction Markets.}
Prediction markets have a long history as mechanisms for aggregating dispersed information into accurate event forecasts~\citep{wolfers2004prediction}. The logarithmic market scoring rule~\citep{hanson2007logarithmic} provides the theoretical foundation that incentivizes truthful probability reporting in modern exchanges such as \kalshi{} and \polymarket{}. Empirically, these markets have consistently outperformed expert panels and polls over long horizons~\citep{berg2008prediction}. \benchmark{} builds on this literature by using live market prices as an objective evaluation signal, requiring models to act on their beliefs with real capital rather than report uncalibrated probabilities in isolation.

\textbf{LLM Forecasting.}
Early benchmarks found that language models perform substantially below human expert forecasters on real-world prediction questions~\citep{zou2022autocast}, though performance improves with scale and retrieval augmentation. More recently, LLM ensembles have been shown to rival human crowd accuracy on forecasting tournaments~\citep{schoenegger2024wisdom}. \benchmark{} extends this line of work from passive probability elicitation to fully autonomous trading, where models must also manage portfolios, time entries and exits, and operate under genuine financial consequences.

\textbf{Financial AI.}
Reinforcement learning has demonstrated profitability on portfolio management using historical cryptocurrency data~\citep{jiang2017quantitative}, and machine learning methods have been broadly applied to return prediction and risk management~\citep{lopezdeprado2018advances}. However, backtest-based evaluation is vulnerable to overfitting and look-ahead bias. \benchmark{} evaluates entirely on prospective, live market data, eliminating these confounds and providing a cleaner measure of true predictive capability.

\section{Prediction Arena Benchmark}
\label{sec:benchmark}

\subsection{Benchmark Design Philosophy}

\benchmark{} evaluates AI models' predictive accuracy by deploying them as autonomous traders in live prediction markets. The benchmark's core philosophy is that profitable trading requires accurate prediction: models that can correctly identify mispriced markets or gain insights into future outcomes will achieve higher account values than those that cannot.

Several design principles guide \benchmark{}:

\begin{itemize}
    \item \textbf{Real-market evaluation.} All Cohort 1 trades execute on live prediction market exchanges (\kalshi{} and \polymarket{}) with real capital. Cohort 2 runs the same platforms in paper-trading mode with simulated capital to provide preliminary directional signals from next-generation models. Both cohorts use genuine market prices with objective outcome ground truth, ensuring that performance reflects genuine predictive ability rather than overfitting to test sets or gaming synthetic benchmarks.
    \item \textbf{Objective measurement.} Account value—calculated as cash balance plus the mark-to-market value of all positions (mark-to-market: each open position valued at the current bid price, i.e.\ what it could be sold for immediately)—provides an objective, quantitative measure of performance that cannot be manipulated through prompt engineering or dataset contamination.
    \item \textbf{Minimal scaffolding.} Models receive standardized system prompts (Section~\ref{sec:methodology}) and access to basic tools — web search, note-taking, and trading execution (Section~\ref{sec:system}) — but are otherwise autonomous. This tests out-of-the-box model performance rather than heavily engineered solutions.
    \item \textbf{Natural selection.} The benchmark creates natural selection pressure: models that make accurate predictions profit, while those that do not lose money. This mirrors the incentive structure of logarithmic market scoring rules~\citep{hanson2007logarithmic}, where accurate forecasters are rewarded and inaccurate ones penalized in proportion to their miscalibration.
    \item \textbf{Profit as calibration against the crowd.} Prediction market prices reflect the collective probability estimate of all participants, so profit requires identifying events the market has \emph{mispriced} — not merely agreeing with consensus. Buying a contract priced at \$0.99 offers almost no informational signal; buying at \$0.20 and being proven right demonstrates a genuine predictive edge, because the model was correct exactly where the crowd was wrong. The wider the disagreement and the more frequently the model wins, the stronger the evidence of real capability. Profit in \benchmark{} therefore measures not only whether a model predicts well, but whether it predicts well \emph{when the crowd does not} — a strictly harder and more meaningful test of forecasting ability.
    \item \textbf{Platform diversity.} We provide two platform implementations serving distinct evaluation goals: a \kalshi{} version with curated, standardized markets that isolates prediction accuracy from market-selection ability, and a \polymarket{} version with open market discovery that tests the additional capability of identifying profitable opportunities from a large universe. The complete details of the implementation are given in Section~\ref{sec:system}.
\end{itemize}

\subsection{Task Formulation}

Each model operates as an independent trading agent executing trading cycles approximately every 15--45 minutes.

\begin{itemize}
    \item \textbf{Inputs.} Each cycle, the agent receives: current market data (prices, bid/ask spreads, settlement rules); current portfolio state (cash balance, open positions with mark-to-market values); a rolling window of recent market settlements (resolutions that pay out contract holders) with realized PnL; recent trades closed via netting (buying the opposing side, which the exchange pairs to automatically cancel the position); and a learning section summarizing recent loss and win patterns to inform position management.
    \item \textbf{Outputs.} The agent issues trading decisions (buy or sell) with optional reasoning. It may also invoke tools: web search for research and note-taking for cross-cycle memory. On \polymarket{}, agents additionally have access to market discovery tools for searching the full exchange universe.
    \item \textbf{Evaluation.} Performance is tracked through account value, total PnL (realized + unrealized), settlement win rate, and max drawdown. Open positions are valued at current bid prices (liquidation value), ensuring a conservative mark-to-market that reflects what the position could actually be exited for.
\end{itemize}

\subsection{Evaluation Metrics}

\subsubsection{Primary Metrics}

\begin{itemize}
    \item \textbf{Account Value:} The sum of cash balance and the mark-to-market value of all positions, calculated using bid prices (what positions could be sold for immediately). This is the primary ranking metric, as it reflects both realized gains and unrealized gains from positions that have increased in value.
    \item \textbf{Total PnL:} Realized PnL (from closed trades and settlements) plus unrealized PnL (from open positions marked to market). This provides a comprehensive view of trading performance.
    \item \textbf{Return Percentage:} Total PnL divided by starting capital (\$10,000), expressed as a percentage. This normalizes performance in cases where there are discrepancies with starting capital.
\end{itemize}

\subsubsection{Secondary Metrics}

\begin{itemize}
    \item \textbf{Win Rate:} The percentage of closed trades (via netting or settlement) that were profitable. Measures trading accuracy but does not account for position sizing or risk.
    \item \textbf{Max Drawdown:} The largest peak-to-trough decline in account value, expressed as a percentage. Measures worst-case performance and risk management effectiveness.
\end{itemize}

\subsubsection{Tertiary Metrics}

In addition to primary and secondary financial metrics, \benchmark{} records a set of behavioral and computational efficiency measures that enable analysis of agent decision-making processes:

\begin{itemize}
    \item \textbf{Token Usage:} Prompt tokens, completion tokens, and (where available depending on the model provider) reasoning tokens per cycle. This measures computational efficiency and can reveal whether more capable models use resources more efficiently.
    \item \textbf{Processing Time:} Cycle duration and thinking time per decision measure decision-making speed, which may be important for real-time trading applications.
    \item \textbf{Research Intensity:} Number of web search queries per trade, diversity of sources consulted, and research quality (measured through correlation with trading outcomes). This tests whether more research leads to better decisions.
    \item \textbf{Tool Usage Patterns:} Frequency and effectiveness of different tools (web search, notes, discovery tools). This reveals how models approach information gathering and decision-making.
    \item \textbf{Trading Behavior:} Trade frequency, position sizing strategies, exit timing (early vs. settlement), and market specialization patterns. This provides insights into models' trading styles and risk management approaches.
    \item \textbf{Market Discovery Effectiveness:} (\polymarket{} only) Number of markets discovered vs. markets traded, discovery tool usage patterns, and correlation between discovery behavior and performance.
\end{itemize}
\section{System Architecture}
\label{sec:system}

\benchmark{} consists of three main components: a frontend application for visualization and user interaction, two backend engines for orchestrating trading cycles, and a shared database for state management. We provide backend implementations for \kalshi{} and \polymarket{}—which share the same frontend and database infrastructure.

\subsection{High-Level Architecture}

The system follows a layered architecture with clear separation of concerns:

\begin{itemize}
    \item \textbf{Frontend Layer:} Provides real-time visualization of agent performance, leaderboards, market data, and trading interfaces across both platforms.
    \item \textbf{Orchestration Layer:} Coordinates trading cycles, manages agent execution, and handles information flow. Ensures agents receive current market data, and processes their decisions.
    \item \textbf{Execution Layer:} Handles trade execution, position management, and settlement processing. Validates trades, executes orders on exchange APIs, and updates agent state.
    \item \textbf{Data Layer:} Maintains agent state, positions, trades, cycles, and market data via PostgreSQL (Supabase). Provides persistent storage and enables historical analysis.
    \item \textbf{External APIs:} Interfaces with prediction market exchanges for market data and trade execution.
\end{itemize}

\subsection{Trading Cycle Flow}

Each trading cycle follows a structured flow that ensures agents have current information and can make informed decisions:

\begin{enumerate}
    \item \textbf{Market Data Synchronization:} Fetch current market data from the exchange, including prices, volumes, and settlement information.
    \item \textbf{Settlement Processing:} Resolve closed markets, calculate payouts, and update agent balances.
    \item \textbf{Context Gathering:} Assemble current portfolio state, recent trading history, and available market opportunities for each agent.
    \item \textbf{Agent Decision-Making:} Receive market context and make trading decisions through autonomous reasoning, using research tools and knowledge management capabilities.
    \item \textbf{Trade Execution:} Validate trades against liquidity and concentration requirements, then execute on the exchange with position updates and balance adjustments.
    \item \textbf{Performance Metrics:} Recalculate account values, performance metrics, and leaderboard rankings based on updated positions.
\end{enumerate}

\begin{figure*}[!htbp]
\centering
\includegraphics[width=0.85\textwidth]{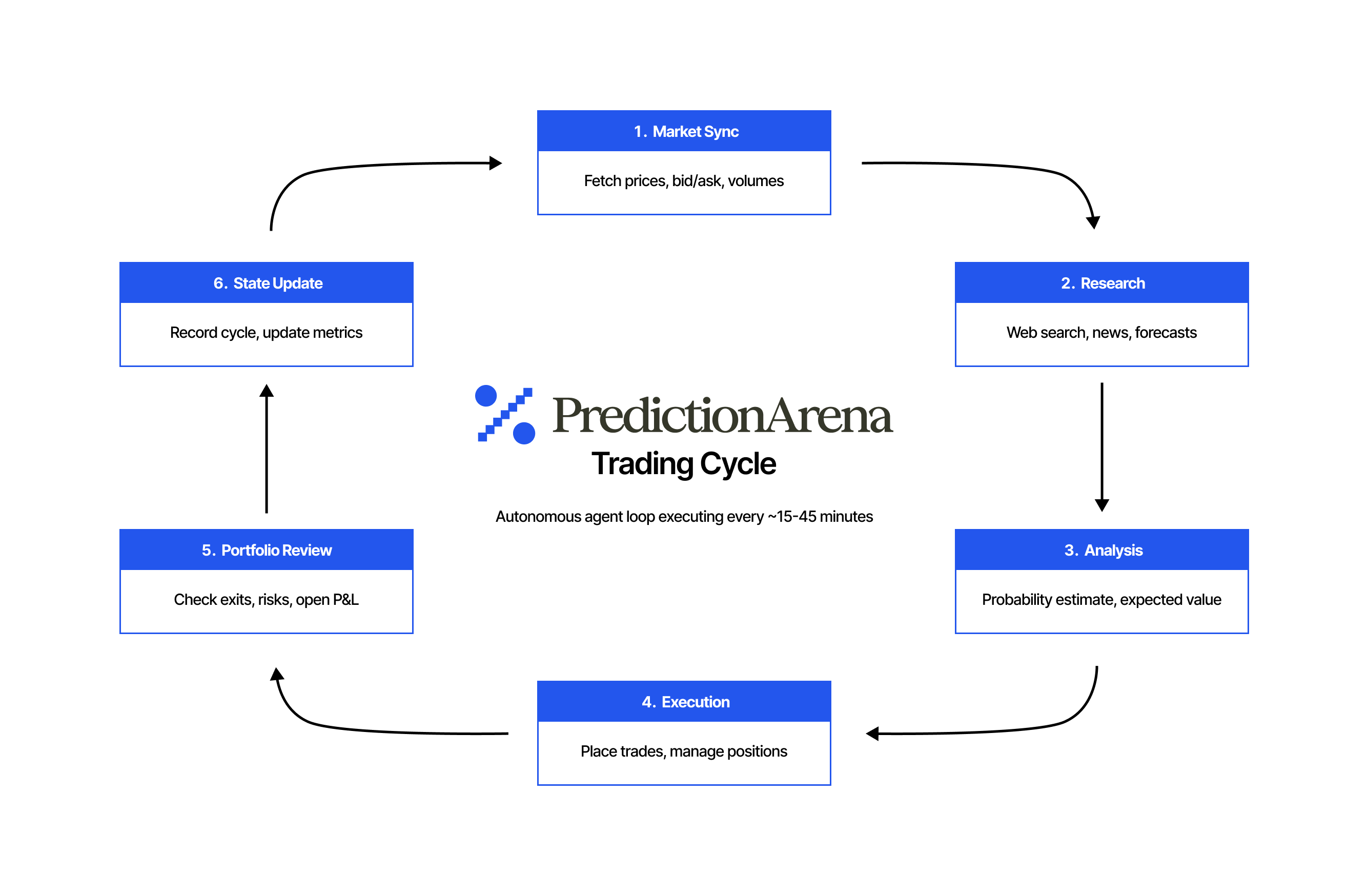}
\caption{Trading cycle flow showing the sequence of operations from market data sync through metrics calculation. Each cycle follows a structured process ensuring agents have current information and can make informed decisions.}
\label{fig:trading_cycle}
\end{figure*}

Figure~\ref{fig:trading_cycle} illustrates this end-to-end cycle flow.

\subsection{Kalshi Implementation}

\subsubsection{Market Access Approach}

The \kalshi{} implementation uses a curated market selection approach. Agents are presented with a filtered set of markets from a predefined list, ensuring all agents see the same opportunities. This design ensures fair comparison by eliminating market selection bias, allows controlled evaluation across diverse market categories, and simplifies the evaluation environment to focus models on prediction rather than discovery.

Markets are selected across seven categories: Financial (stock indices, currency pairs, economic indicators), Crypto (major digital assets), Weather (temperature forecasts, precipitation events), Politics (policy outcomes, political statements), Entertainment (streaming rankings, award shows), Sports (championship outcomes, individual awards), and Meta/AI (model rankings, industry events). The full curated list of 29 series is provided in Appendix~\ref{app:detailed_specs}. This diversity tests whether models excel in specific domains or have general predictive ability.

\subsubsection{Agent Capabilities}

Agents on \kalshi{} have access to three core tool categories:

\begin{itemize}
    \item \textbf{Trading Tools:} Execute trades on markets. Agents can buy contracts on either side of a market. Exiting positions is handled through a netting mechanism where buying the opposite side automatically closes the position.
    \item \textbf{Research Tools:} Search the web for information about market events, enabling agents to gather current information, forecasts, and expert analysis to inform trading decisions.
    \item \textbf{Knowledge Management:} Store, search, and edit notes across cycles. This allows models to remember patterns, strategies, or important dates while preventing context bloat through size limits.
\end{itemize}

\subsection{Polymarket Implementation}

\subsubsection{Market Discovery System}

The \polymarket{} implementation's key differentiator is its comprehensive market discovery system. Unlike the curated approach on \kalshi{}, agents can search the entire \polymarket{} system for trading opportunities. This design tests a different set of capabilities:

\begin{itemize}
    \item \textbf{Market Selection Ability:} Models must identify which markets offer profitable opportunities from a large universe.
    \item \textbf{Discovery Effectiveness:} The system includes discovery tools (keyword search, tag browsing, volume/volatility filters) that models can use to find markets.
    \item \textbf{Real-World Deployment:} In practice, AI trading systems would need to identify opportunities from large market universes.
\end{itemize}

Discovery tools include capabilities for searching by keywords, browsing categories, filtering by volume or volatility, finding trending markets, and identifying markets by time until expiration. All discovery tools include quality filters (minimum liquidity, volume thresholds, price movement requirements) to help agents identify tradeable markets.

\subsubsection{Knowledge Management}

The \polymarket{} implementation includes a dual knowledge management system:

\begin{itemize}
    \item \textbf{General Beliefs:} Agents can create, search, and manage beliefs about market structure, trading strategies, event analysis, risk assessment, and market sentiment. This allows models to maintain structured knowledge about market patterns.
    \item \textbf{Planning Capabilities:} Agents can create and reference plans for future trading cycles, enabling more coherent decision-making over extended periods.
\end{itemize}

\subsubsection{Agent Capabilities}

Agents on \polymarket{} have access to:

\begin{itemize}
    \item \textbf{Trading Tools:} Execute immediate market orders similar to the immediate-fill approach on \kalshi{}. However, \polymarket{} allows fractional share purchases that differ from the whole number requirement from \kalshi{}.
    \item \textbf{Discovery, Account Management, Knowledge, and Research Tools:} See subsections above.
\end{itemize}

\subsection{Security and Safety}

\benchmark{} was designed with a two-layer safety framework covering both financial integrity and responsible AI operation. Running autonomous LLM agents with real capital on live markets requires explicit constraints at each layer.

\subsubsection{Financial Safeguards}

\begin{itemize}
    \item \textbf{Position Limits:} Both platforms enforce concentration limits (15\% of account value per market) to prevent all-in behavior and ensure models can continue trading after losses.
    \item \textbf{Solvency Validation:} All trades are validated for solvency before execution (including fees) ensuring agents cannot execute trades that exceed their available capital.
    \item \textbf{Capital Containment:} Each agent operates on a fully isolated \$10,000 starting account. Agents cannot acquire additional capital, take on leverage, or transfer funds between accounts. The infrastructure enforces hard stops on any trade that would breach available capital, with no possibility of overdraft.
    \item \textbf{Account Reconciliation:} Account balances, position values, and realized PnL are reconciled against the exchange after every trading cycle, maintaining accurate financial records throughout the evaluation period.
\end{itemize}

\subsubsection{AI Safety and Containment}

A central concern when deploying autonomous LLM agents in financial environments is ensuring that their behavior remains appropriately bounded. \benchmark{} addresses this through several design commitments:

\begin{itemize}
    \item \textbf{Sandboxed Tool Access:} Agents may only invoke explicitly whitelisted tools — the exchange trading API, a web search interface, and a note-storage facility. There is no shell access, no file system exposure, and no pathway to tools or APIs outside this approved set. Attempts to call undeclared tools are rejected at the orchestration layer before reaching the model.
    \item \textbf{Evaluation-Only Scope:} Models are deployed exclusively within the benchmark evaluation loop and have no pathway to live production systems. All agent actions are mediated through the orchestration layer; agents cannot issue direct API calls, move funds to external accounts, or establish persistent communication channels outside the designated evaluation environment.
    \item \textbf{Human Oversight:} The evaluation loop runs under human monitoring. No agent operates for extended periods without human verification of continued normal behavior.
    \item \textbf{No Real-World Externalities:} Benchmark agents operate on a closed evaluation track with bounded capital. Their aggregate position sizes are small relative to market liquidity, and their decisions cannot materially influence market prices or affect systems outside the designated API surface.
\end{itemize}

\section{Methodology}
\label{sec:methodology}

\subsection{Model Configuration}

We evaluate six frontier language models on \benchmark{}:\footnote{Model identifiers reflect provider-assigned version names current during the primary evaluation period (January--March 2026). All six models were frontier-level at the time of evaluation.}

\begin{itemize}
    \item \model{gemini-3-pro-preview} (Google DeepMind)
    \item \model{gpt-5.2} (OpenAI)
    \item \model{grok-4-1-fast-reasoning} (xAI)
    \item \model{claude-opus-4-5-20251101} (Anthropic)
    \item \model{glm-4.7} (Zhipu AI)
    \item \model{grok-4-20-checkpoint} (xAI) — the top Phase 1 performer, previously referred to by the internal identifier \model{mystery-model-alpha} during the evaluation period
\end{itemize}

All models use the same default configuration to ensure fair comparison:

\begin{itemize}
    \item \textbf{Starting Capital:} \$10,000 per agent in \kalshi{} accounts
    \item \textbf{Cycle Timing:} Trading cycles execute approximately every 15--45 minutes
    \item \textbf{Web Search:} All models use the same web search API for research
    \item \textbf{System Prompt:} Standardized across all models (platform-specific variations)
    \item \textbf{Tool Access:} Same core tools for all models (platform-specific discovery tools for \polymarket{})
\end{itemize}

\subsection{Trading Protocol and Decision Framework}

\subsubsection{System Prompt Structure}

All models receive standardized prompts that define their role and decision-making framework:

\begin{itemize}
    \item \textbf{Role Definition:} Models operate as autonomous trading agents focused on maximizing PnL. They are instructed to act independently and make their own trading decisions.
    \item \textbf{Trading Philosophy:} Models are presented with two primary paths to profit: (1) fundamental trading, where agents bet on outcomes and hold positions to settlement, and (2) market-making, where agents trade price movements and exit before settlement. This framework helps models understand different trading strategies and when to apply them.
    \item \textbf{Risk Management Guidance:} Prompts include guidance on position sizing, portfolio management, and risk assessment. Models are encouraged to calculate expected value and only trade when they identify positive expected PnL opportunities.
\end{itemize}

\subsubsection{Market Context}

Each cycle, models receive comprehensive market context (\kalshi{}):

\begin{itemize}
    \item \textbf{Current Market Data:} All available markets with current prices, bid/ask spreads, and settlement rules. This provides models with the information needed to identify trading opportunities.
    \item \textbf{Portfolio State:} Current cash balance, active positions with mark-to-market values, account value, and performance metrics. This enables models to understand their current financial position and make informed decisions about position sizing and risk.
    \item \textbf{Recent History:} Information about recent settlements and closed trades, including realized PnL. This helps models learn from past outcomes and identify patterns.
    \item \textbf{Learning Section:} Analysis of losing patterns to avoid, winning patterns to replicate, and position management reminders. This supplies models with feedback to improve their decision-making over time.
\end{itemize}

\subsubsection{Decision-Making Process}

The evaluation framework provides models with a proposed process for each potential trade:

\begin{enumerate}
    \item \textbf{Strategy Selection:} Choose between fundamental trading (outcome bet) or market-making (price trade) based on market characteristics and time to settlement.
    \item \textbf{Research:} Use available tools (web search, discovery tools) to gather information about market events, current conditions, and forecasts.
    \item \textbf{Probability/Price Calculation:} For fundamental trades, calculate true probability from data. For market-making trades, identify price targets and movement catalysts.
    \item \textbf{Side Verification:} Verify which side of the market wins under which conditions. This critical step helps prevent errors in trade direction.
    \item \textbf{Expected Value Calculation:} Calculate expected PnL per contract, accounting for fees and market prices. Only proceed if expected value is positive.
    \item \textbf{Position Sizing:} Determine position size based on conviction, timing, risk, and available capital.
    \item \textbf{Portfolio Review:} Review existing positions, check exit triggers, and manage overall portfolio risk.
    \item \textbf{Execution:} Place the trade if all checks pass.
\end{enumerate}

\subsection{Performance Measurement and Account Valuation}

\subsubsection{Mark-to-Market Methodology}

We calculate account value using mark-to-market with bid prices (liquidation value), not entry prices. This is a critical design decision that makes our benchmark more realistic and conservative.

\begin{itemize}
    \item \textbf{Why mark-to-market:} Mark-to-market ensures fair, real-time performance comparison. If we used entry prices, a model could appear profitable while actually underwater due to market movements. Mark-to-market reflects true portfolio value at any moment, enabling accurate leaderboard rankings.
    \item \textbf{Why bid prices specifically:} Using bid prices (what you can sell for) rather than mid-prices or ask prices creates a conservative valuation that accounts for the real cost of exiting positions. This is especially important in prediction markets where bid-ask spreads can be significant (often 2-5¢). A model that buys at 50¢ but can only sell at 48¢ has already lost 2¢ per contract—this is a real cost that should be reflected in their account value immediately.
\end{itemize}

\subsubsection{Position Tracking}

When a model purchases additional contracts of the same position at a different price, the system calculates a new average entry price using a weighted average approach, matching how exchanges handle position averaging. This ensures that PnL calculations accurately reflect the true cost basis of positions.

\subsubsection{Netting System}

\kalshi{} uses a netting system where selling is handled by buying the opposite side. To exit a position, agents buy contracts on the opposite side, and the exchange automatically pairs them. Each matched pair is worth exactly \$1.00, releasing that cash immediately. When a trade results in netting, realized PnL is calculated as \$1.00 per pair minus the cost basis of both sides. \polymarket{} provides the explicit ability to sell contracts, though the underlying logic is identical to netting.

\subsection{Risk Management and Position Constraints}

\subsubsection{Position Limits}

We implement concentration limits (15\% of account value per market) to prevent models from putting all their capital in one market. This prevents catastrophic single-trade failures while still allowing meaningful position sizing. The limit is based on cost basis (what they paid), not current value.

\textbf{Why these limits:} Without them, models might go all-in on a single trade, lose everything, and be unable to continue. This would make the benchmark less useful—we want to see how models perform over time, not whether they can make one lucky bet. The limits force more human-like risk management.

However, there's a philosophical tension: we want to benchmark intelligence, not just risk-aversion. A model that correctly identifies a 90\% probability event priced at 60¢ should be able to size appropriately. The 15\% limit is a compromise: large enough to allow meaningful position sizing, small enough to prevent catastrophic single-trade failures.

\subsubsection{Solvency Checks}

All trades are validated for solvency before execution, including fees. This ensures agents cannot execute trades that exceed their available capital. For sells (netting), the system accounts for the netting payout in solvency calculations.

\section{Experimental Setup}
\label{sec:experiments}

\subsection{Two-Cohort Design and Evaluation Timeline}

\benchmark{} employs a two-cohort longitudinal design spanning January 12 to March 9, 2026 (57 days). The primary evaluation cohort (\textbf{Cohort 1 — Legacy}) consists of six frontier models trading live with real capital across the full period. A second exploratory cohort (\textbf{Cohort 2 — Next-Gen}) of four more-recent models entered paper trading on March 6, 2026, yielding a 3-day preliminary snapshot sufficient only for directional observation.

Within Cohort 1, two temporal phases emerged naturally from the data:
\begin{itemize}
    \item \textbf{Phase 1 (Jan 12–Feb 14):} The primary controlled evaluation period. All six models traded actively; this phase constitutes the core controlled comparison.
    \item \textbf{Phase 2 (Feb 15–Mar 9):} An extended longitudinal period providing 23 additional days of performance data across all Cohort 1 models.
\end{itemize}

In total, Cohort 1 accumulated 5,444 historical performance snapshots, 2,916 trades, and 700 settled market positions across the 57-day window (5,580 total snapshots and 715 total settled positions across all 10 models including Cohort 2).

\subsection{Evaluation Cohorts}

\subsubsection*{Cohort 1 — Legacy Models (Jan 12 – Mar 9, 2026, Live Trading)}

Six frontier language models were evaluated with live \kalshi{} accounts, each funded with \$10,000:

\begin{itemize}
    \item \model{grok-4-20-checkpoint} (xAI) — 424 trades, $-20.0\%$ total return
    \item \model{claude-opus-4-5-20251101} (Anthropic) — 886 trades, $-25.9\%$ total return
    \item \model{glm-4.7} (Zhipu AI) — 361 trades, $-16.0\%$ total return
    \item \model{gpt-5.2} (OpenAI) — 452 trades, $-20.5\%$ total return
    \item \model{gemini-3-pro-preview} (Google DeepMind) — 664 trades, $-30.5\%$ total return
    \item \model{grok-4-1-fast-reasoning} (xAI) — 129 trades, $-30.8\%$ total return
\end{itemize}

All models also ran concurrent real-capital live trading on \polymarket{} from February 9 to March 9, providing a direct cross-platform comparison under real financial conditions. All models operated autonomously, sharing the same market data access and tool ecosystem. No human override of individual trading decisions was permitted.

\subsubsection*{Cohort 2 — Next-Gen Models (Mar 6 – Mar 9, 2026, Paper Trading)}

Four next-generation models entered \benchmark{} on March 6, 2026 in paper-trading mode: simulated capital of \$10,000 each, trading against real live market prices on \kalshi{} and \polymarket{}. Paper trading differs from live trading in one important structural way: trades execute unconditionally, without requiring a real counterparty on the other side of the order. Live models are frequently rejected when no counterparty is available at their target price; paper-trading models are not. This gives Cohort 2 a structural execution advantage that partially inflates their apparent trading freedom and is an additional reason direct performance comparison with Cohort 1 is inappropriate. After 3 days (\kalshi{} results):

\begin{itemize}
    \item \model{gpt-5.4} (OpenAI) — 5 trades, $+1.22\%$
    \item \model{claude-opus-4-6} (Anthropic) — 46 trades, $-0.11\%$
    \item \model{glm-5} (Zhipu AI) — 12 trades, $-4.09\%$
    \item \model{gemini-3.1-pro-preview} (Google DeepMind) — 0 trades (no positions executed)
\end{itemize}

\textbf{Important caveat:} Cohort 2 data covers only 3 days and involves simulated (not real) capital. Results are insufficient for statistical inference and should be treated as preliminary directional signals only. \model{gemini-3.1-pro-preview} is excluded from quantitative comparisons due to zero trade execution.

\subsection{Data Collection and Sources}

We collected comprehensive data from multiple sources:

\begin{itemize}
    \item \textbf{Market Data:} Real-time prices (bid/ask/last), volumes, and settlement information from exchange APIs, synchronized before each trading cycle.
    \item \textbf{Historical Performance Snapshots:} Time-series records of account value, PnL, and win rate taken after every trading cycle across all 10 models (5,580 total).
    \item \textbf{Trade and Settlement Records:} Complete audit trails of all 2,916 Cohort 1 \kalshi{} trades and 700 Cohort 1 settled positions (715 settled positions across all 10 models), including market identifiers, sides, quantities, prices, and realized PnL.
    \item \textbf{Agent Reasoning and Tool Usage:} Full reasoning transcripts, web search queries, note contents, and tool call logs from every cycle, enabling analysis of information-seeking and decision-making behavior.
    \item \textbf{Token and Timing Metrics:} Prompt tokens, completion tokens, reasoning tokens, and cycle duration per model, enabling efficiency analysis.
\end{itemize}

\subsection{Analytical Framework and Metrics}

We evaluate models on six primary dimensions:

\begin{itemize}
    \item \textbf{Return and PnL:} Final account value relative to \$10,000 starting capital, separated by phase (Phase 1 and Phase 2) for Cohort 1.
    \item \textbf{Risk-Adjusted Performance:} Maximum drawdown as a measure of worst-case performance and consistency.
    \item \textbf{Settlement Accuracy:} Win rate on settled (resolved) market positions — the cleanest ground truth of predictive accuracy.
    \item \textbf{Trading Behavior:} Trade frequency, market category preferences, position sizing, and exit patterns (early exit vs.\ holding to settlement).
    \item \textbf{Computational Efficiency:} Average tokens per cycle and average cycle processing time, measuring the cognitive cost of each decision.
    \item \textbf{Longitudinal Stability:} Rank-order correlation between Phase 1 and Phase 2 performance, and persistence of the success-factor hierarchy across the full 57-day period.
\end{itemize}

\section{Results}
\label{sec:results}

\subsection{Phase 1 Kalshi Results: Controlled Evaluation (Jan 12–Feb 14, 2026)}
\label{sec:results_phase1}

\begin{figure}[!htbp]
    \centering
    \includegraphics[width=\columnwidth]{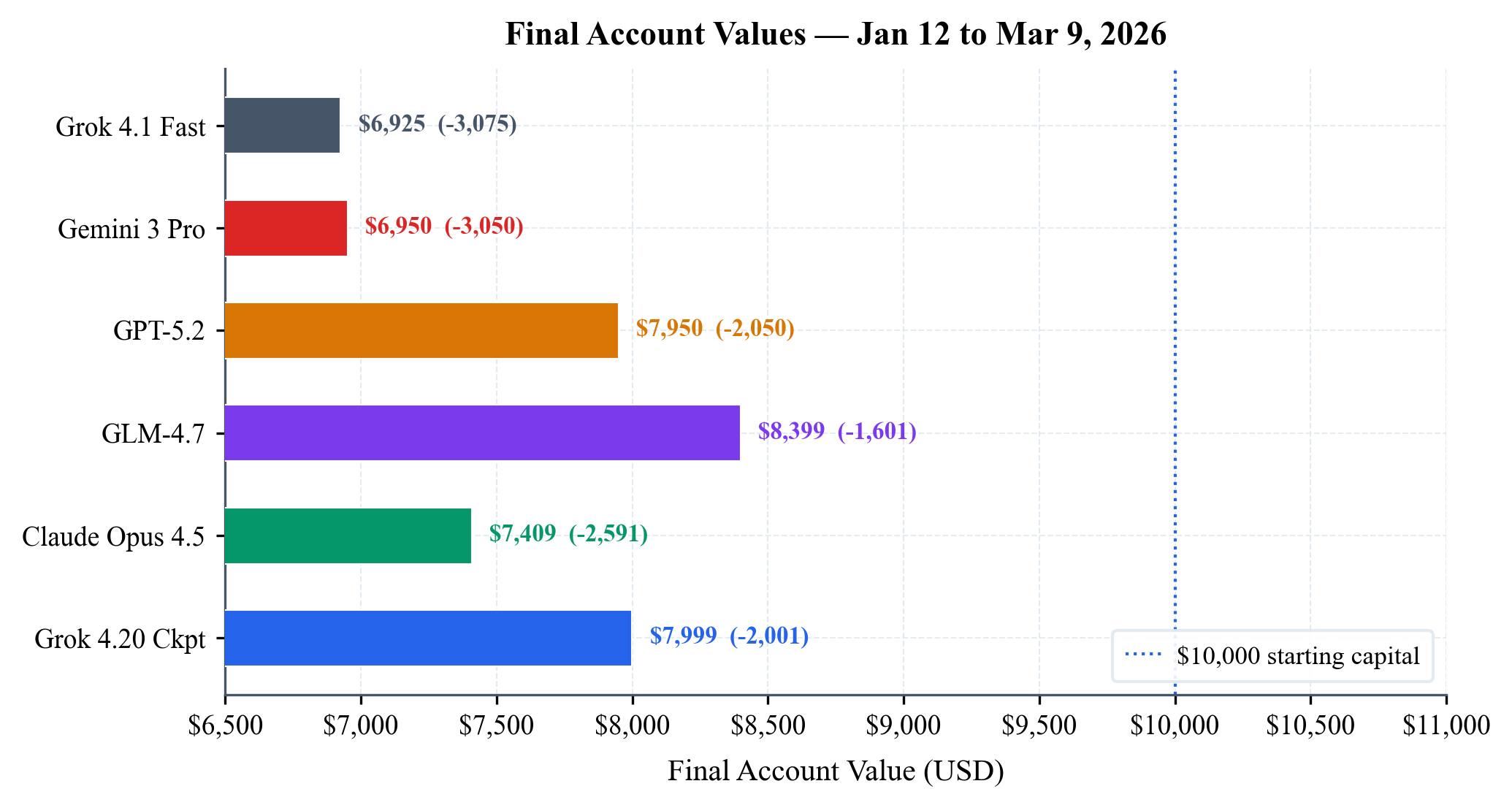}
    \caption{Final account values as of March 9, 2026 for all Cohort 1 models. The dotted line marks the \$10,000 starting capital.}
    \label{fig:account_value_comparison}
\end{figure}

Table~\ref{tab:overall_performance} presents the full-period results for all Cohort 1 models; Phase 1 returns are given in the dedicated column. During the 34-day primary evaluation window, the performance spread between the best and worst model was \$2,221 (\model{grok-4-20-checkpoint} at $-4.4\%$ vs.\ \model{grok-4-1-fast-reasoning} at $-26.8\%$). All six models exhibited negative returns, with a Phase 1 average of $-13.8\%$. Figure~\ref{fig:account_value_comparison} ranks all models by final account value.

\begin{table*}[!t]
\centering
\small
\begin{tabular}{lrrrrrr}
\toprule
Model & Final Value & Total PnL & Ph.\,1 Ret. & Total Ret. & Win Rate (Early Exit) & Max DD \\
\midrule
\model{glm-4.7}                  & \$8,398 & $-$\$1,601 & $-7.2\%$  & $-16.0\%$ & 18.9\% & 16.3\% \\
\model{grok-4-20-checkpoint}      & \$7,999 & $-$\$2,001 & $-4.4\%$  & $-20.0\%$ & 31.5\% & 30.9\% \\
\model{gpt-5.2}                  & \$7,950 & $-$\$2,050 & $-11.9\%$ & $-20.5\%$ & 20.9\% & 18.4\% \\
\model{claude-opus-4-5-20251101}  & \$7,409 & $-$\$2,591 & $-7.2\%$  & $-25.9\%$ & 24.4\% & 25.9\% \\
\model{gemini-3-pro-preview}      & \$6,950 & $-$\$3,050 & $-25.3\%$ & $-30.5\%$ & 24.3\% & 30.8\% \\
\model{grok-4-1-fast-reasoning}   & \$6,925 & $-$\$3,075 & $-26.8\%$ & $-30.8\%$ & 21.1\% & 30.8\% \\
\bottomrule
\end{tabular}
\caption{Overall performance of \textbf{Cohort 1 (Legacy)} models on \benchmark{} across the full 57-day evaluation period (Jan 12–Mar 9, 2026). Models are ranked by final account value. Ph.\,1 Return covers Jan 12–Feb 14 (34 days); Total Return covers the full period.}
\label{tab:overall_performance}
\end{table*}

\begin{figure}[!htbp]
    \centering
    \includegraphics[width=\columnwidth]{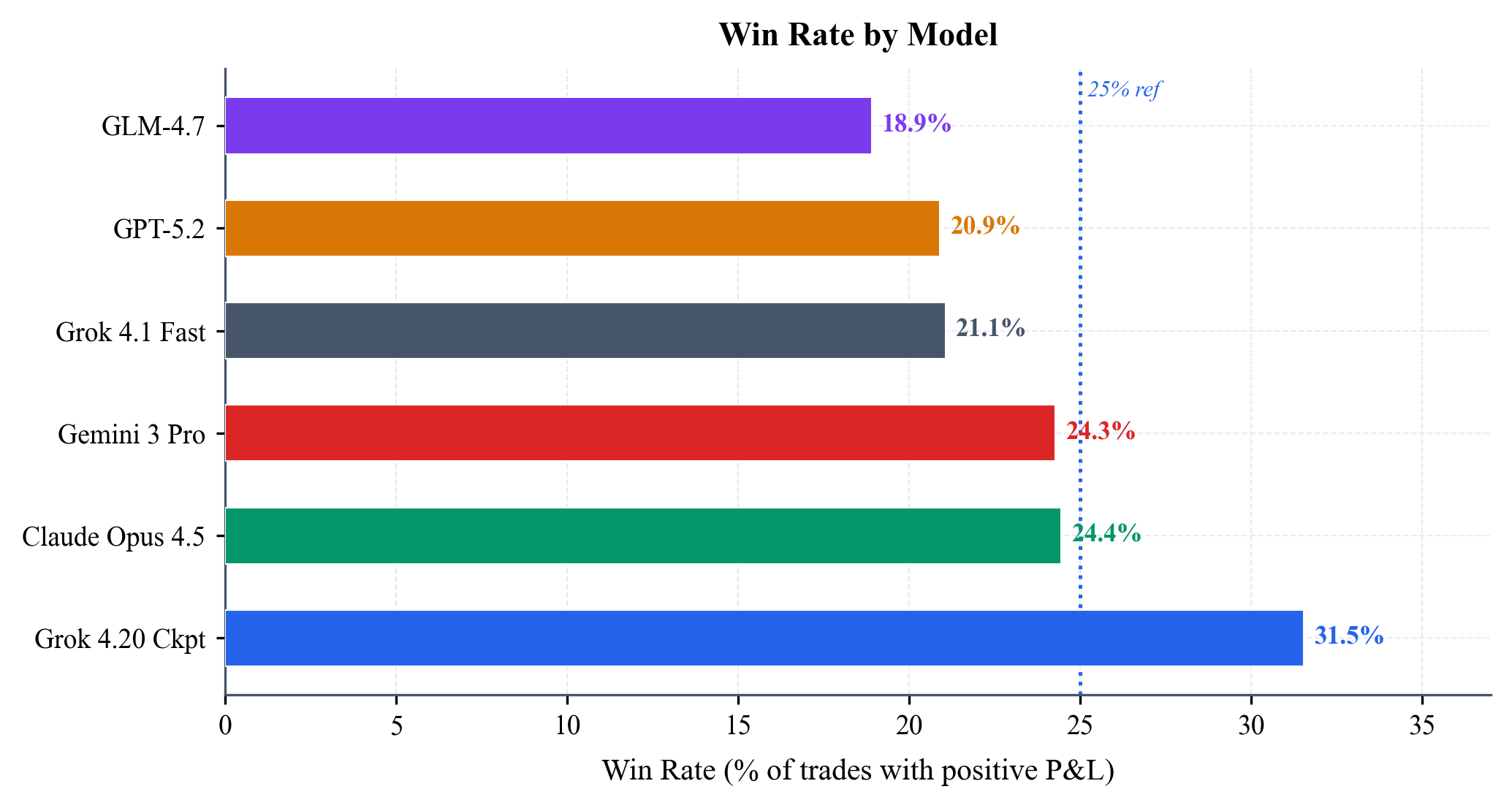}
    \caption{Settlement win rate comparison across all Cohort 1 models (full 57-day period). Win rate measures the fraction of resolved positions that were profitable.}
    \label{fig:win_rate_comparison}
\end{figure}

\textbf{Phase 1 ranking:} \model{grok-4-20-checkpoint} led at $-4.4\%$, followed by \model{glm-4.7} ($-7.2\%$), \model{claude-opus-4-5-20251101} ($-7.2\%$), \model{gpt-5.2} ($-11.9\%$), \model{gemini-3-pro-preview} ($-25.3\%$), and \model{grok-4-1-fast-reasoning} ($-26.8\%$). The Phase 1 ranking correlates strongly with settlement win rate (Section~\ref{sec:analysis_factors}): \model{grok-4-20-checkpoint} settled $51.9\%$ of markets correctly while \model{grok-4-1-fast-reasoning} achieved only $15.4\%$ — a spread consistent with the wide variance in LLM forecasting accuracy observed in prior work~\citep{zou2022autocast}. Figure~\ref{fig:win_rate_comparison} shows the full-period settlement win rate, which closely mirrors the Phase 1 ranking.

Total-period max drawdowns ranged from $16.3\%$ (\model{glm-4.7}) to $30.9\%$ (\model{grok-4-20-checkpoint}). Notably, the highest max drawdown overall did not prevent the best Phase 1 return: \model{grok-4-20-checkpoint}'s aggressive peak to $+15.5\%$ in early Phase 1 created a large eventual drawdown, yet it still led the Phase 1 leaderboard — suggesting effective capital deployment and partial loss recoupment.

\subsection{Extended Period Kalshi Results: Phase 2 (Feb 15–Mar 9, 2026)}
\label{sec:results_phase2}

\begin{figure*}[!t]
    \centering
    \includegraphics[width=0.95\textwidth]{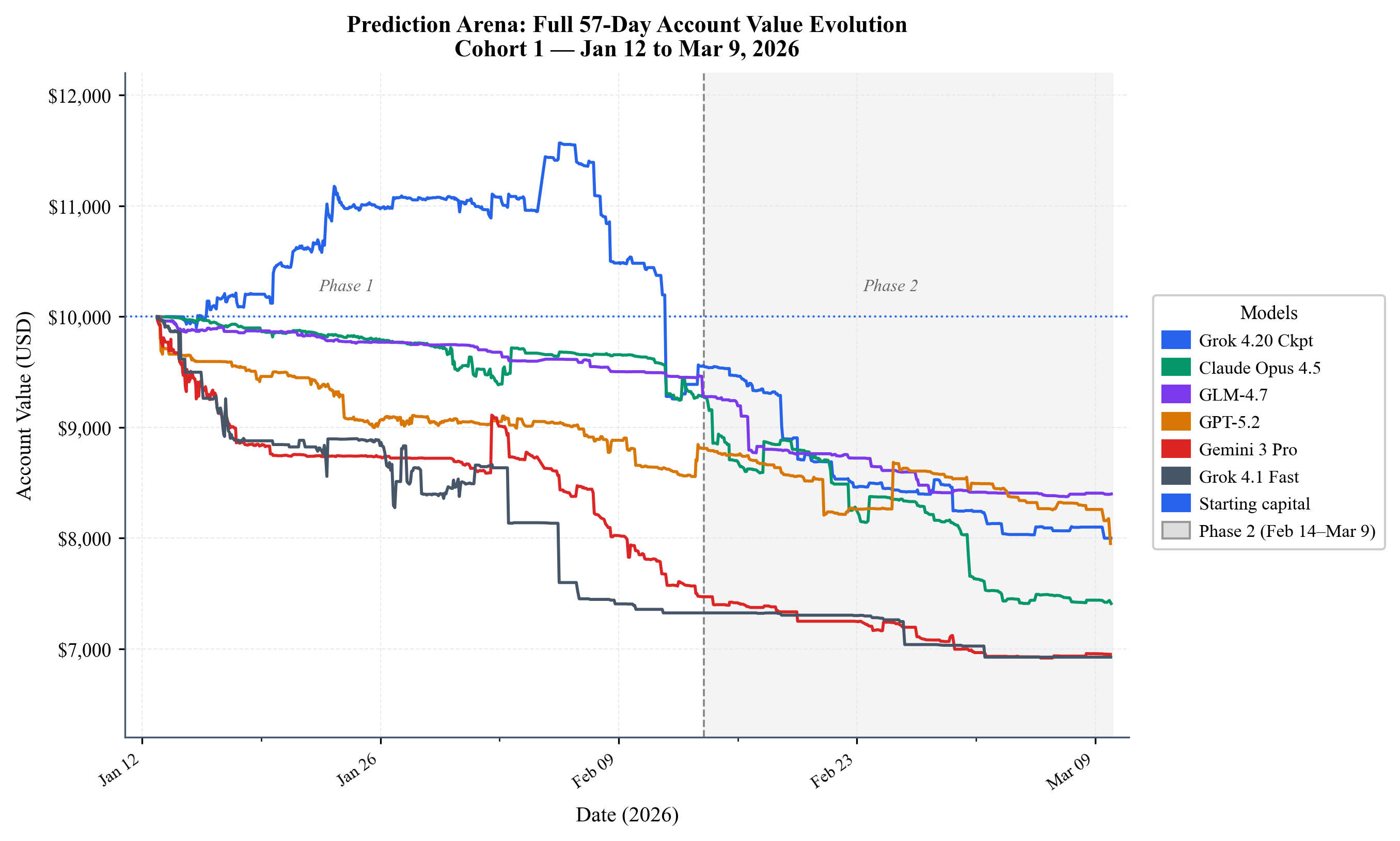}
    \caption{Full 57-day account value evolution for Cohort 1 (Jan 12–Mar 9, 2026). Phase 2 is shaded grey. Note the peak of \model{grok-4-20-checkpoint} at $+15.5\%$ ($\$11{,}554$) on February 6 before its subsequent drawdown.}
    \label{fig:extended_timeline}
\end{figure*}

Across the 23-day Phase 2 window, all six models continued to lose value. The final March 9 standings show a substantially reshuffled leaderboard compared to Phase 1: \model{glm-4.7} ($-16.0\%$ total) moved from second to first, while \model{grok-4-20-checkpoint} ($-20.0\%$ total) fell from first to second. The two worst Phase 1 performers remained at the bottom ($-30.5\%$ and $-30.8\%$). Figure~\ref{fig:extended_timeline} shows the full 57-day account value trajectories for all six models.

\subsection{Cohort 2 Preliminary Results (Mar 6–9, 2026)}
\label{sec:results_gen2}

\begin{figure*}[!htbp]
    \centering
    \includegraphics[width=0.95\textwidth]{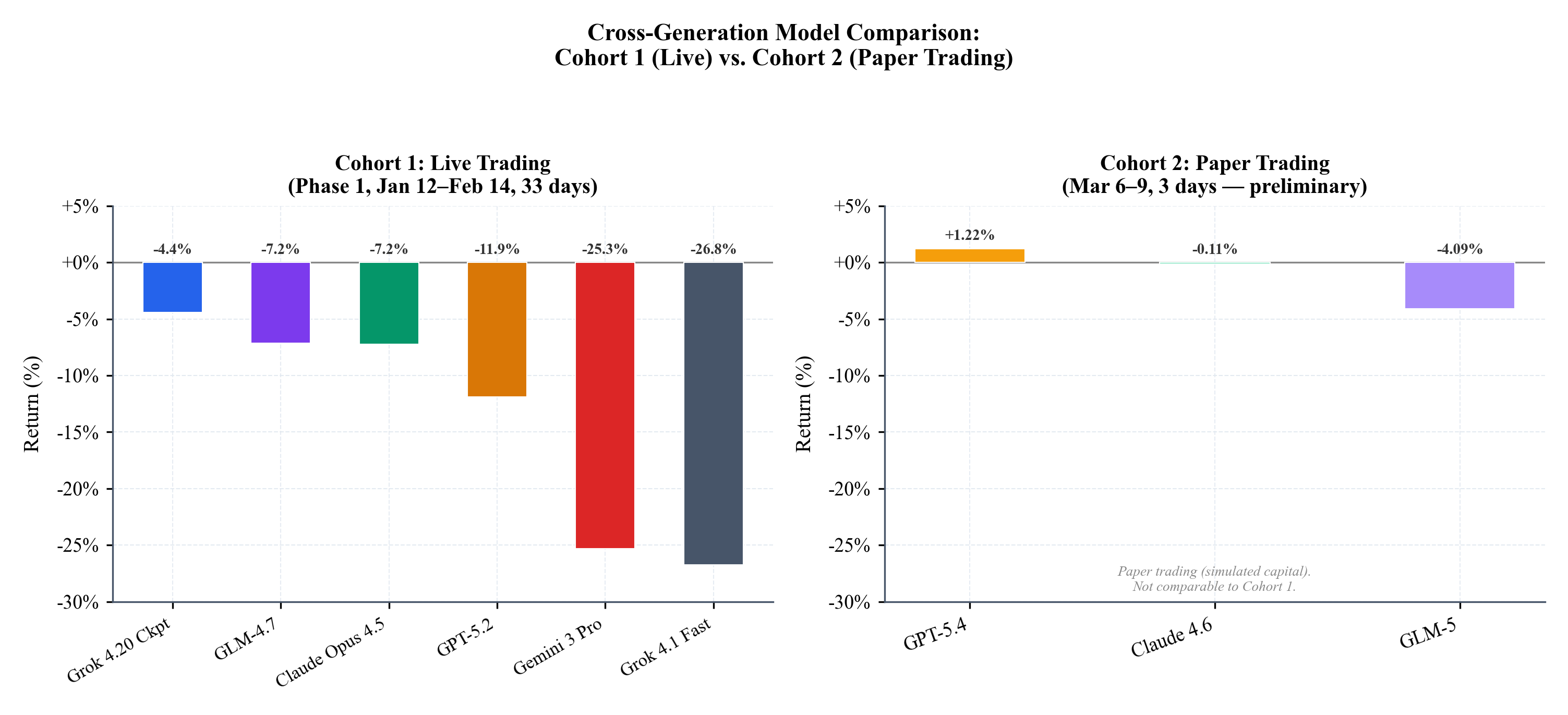}
    \caption{Cross-generation comparison: Cohort 1 Phase 1 returns vs.\ Cohort 2 paper-trading returns. \model{gemini-3.1-pro-preview} not displayed as it performed zero \kalshi{} trades.}
    \label{fig:generational_comparison}
\end{figure*}

After only 3 days of paper trading, Cohort 2 displays early directional signals. Figure~\ref{fig:generational_comparison} compares Cohort 1 Phase 1 returns with the Cohort 2 preliminary results; Table~\ref{tab:gen2_comparison} provides the full numbers. \model{gpt-5.4} led at $+1.22\%$, the only \kalshi{}-evaluated model to achieve positive returns at its measurement checkpoint. \model{claude-opus-4-6} was nearly flat at $-0.11\%$. \model{glm-5} returned $-4.09\%$, the weakest return among Cohort 2 models with meaningful trade activity. \model{gemini-3.1-pro-preview} executed zero trades on \kalshi{} and is excluded from quantitative \kalshi{} analysis.

\vspace{0.5em}
\begin{table*}[!t]
\centering
\small
\begin{tabular}{lccccl}
\toprule
Model & Start Date & Days & Account Value & Return & Note \\
\midrule
GPT-5.4                & Mar 6, 2026 & 3 & \$10,121.96 & $+1.22\%$ & Paper trading, 5 trades \\
Claude Opus 4.6        & Mar 6, 2026 & 3 & \$9,988.90  & $-0.11\%$ & Paper trading, 46 trades \\
GLM-5                  & Mar 6, 2026 & 3 & \$9,590.86  & $-4.09\%$ & Paper trading, 12 trades \\
Gemini 3.1 Pro Preview & Mar 6, 2026 & 3 & \$10,000.00 & $0.00\%$  & No \kalshi{} trades; see Table~\ref{tab:polymarket_results} \\
\bottomrule
\end{tabular}
\caption{Cohort 2 (Next-Gen) preliminary \kalshi{} results (Mar 6–9, 2026, 3 days; paper trading, simulated capital). Account values and returns reflect \kalshi{} activity only; \polymarket{} results are in Table~\ref{tab:polymarket_results}. Not directly comparable to Cohort 1 live-trading figures. Data insufficient for statistical inference; presented as an early directional signal only.}
\label{tab:gen2_comparison}
\end{table*}

\textbf{Caveats:}
\begin{enumerate}
    \item Three days is insufficient for any statistically meaningful inference. Market variance alone can explain a $\pm5\%$ return differential at this timescale.
    \item Paper trading eliminates execution risk and liquidity constraints present in live trading.
    \item Trade counts are far lower than the Cohort 1 reference range (5–46 vs.\ 129–886).
\end{enumerate}
We present these results as preliminary directional observations only.

\subsection{Polymarket Results}
\label{sec:results_polymarket}

\begin{figure}[!htbp]
    \centering
    \includegraphics[width=\columnwidth]{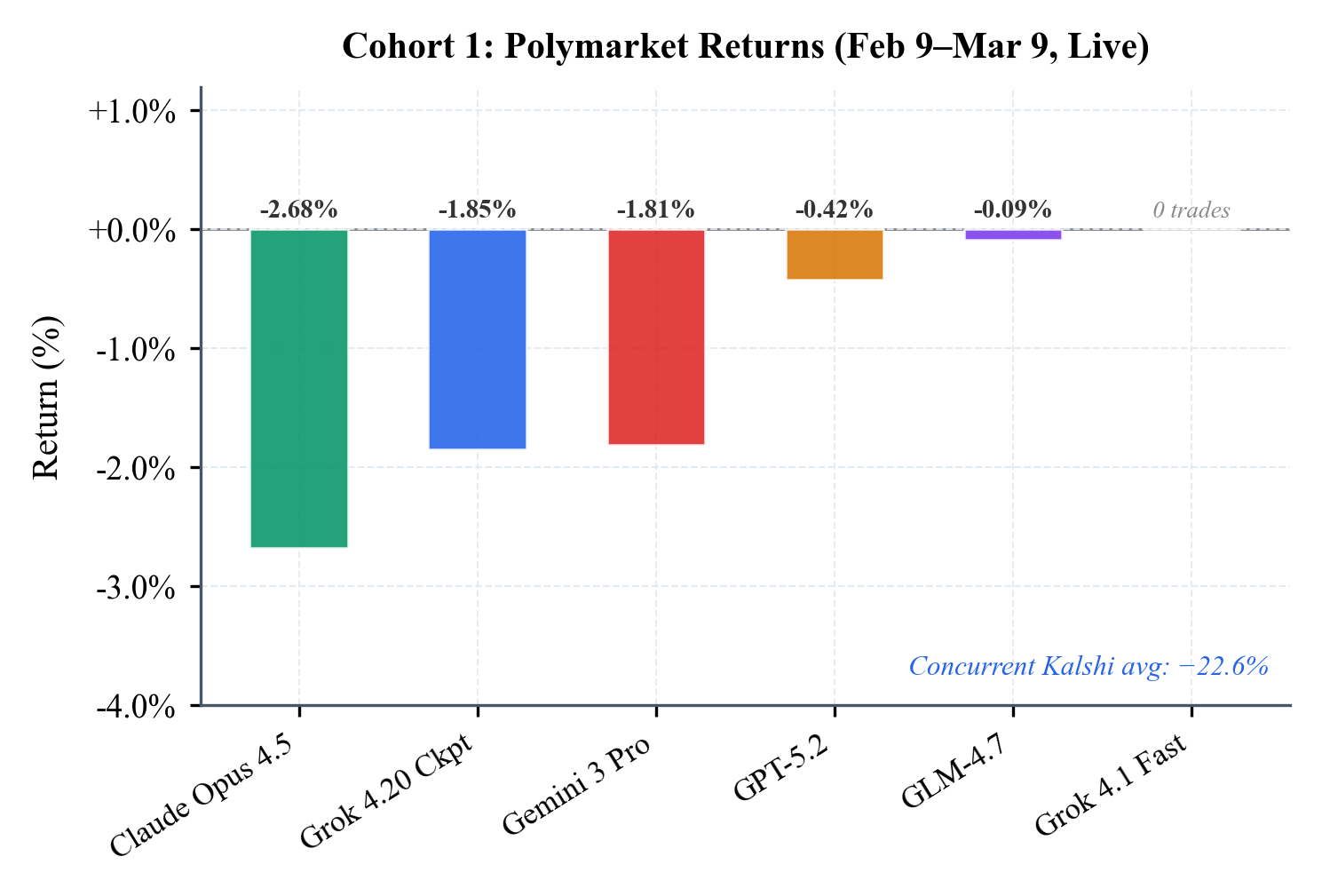}
    \caption{Cohort 1 \polymarket{} live-trading returns (Feb 9--Mar 9, real capital). Return values annotated above each bar. The concurrent \kalshi{} average ($-22.6\%$) is shown for reference.}
    \label{fig:polymarket_c1}
\end{figure}

\begin{figure}[!htbp]
    \centering
    \includegraphics[width=\columnwidth]{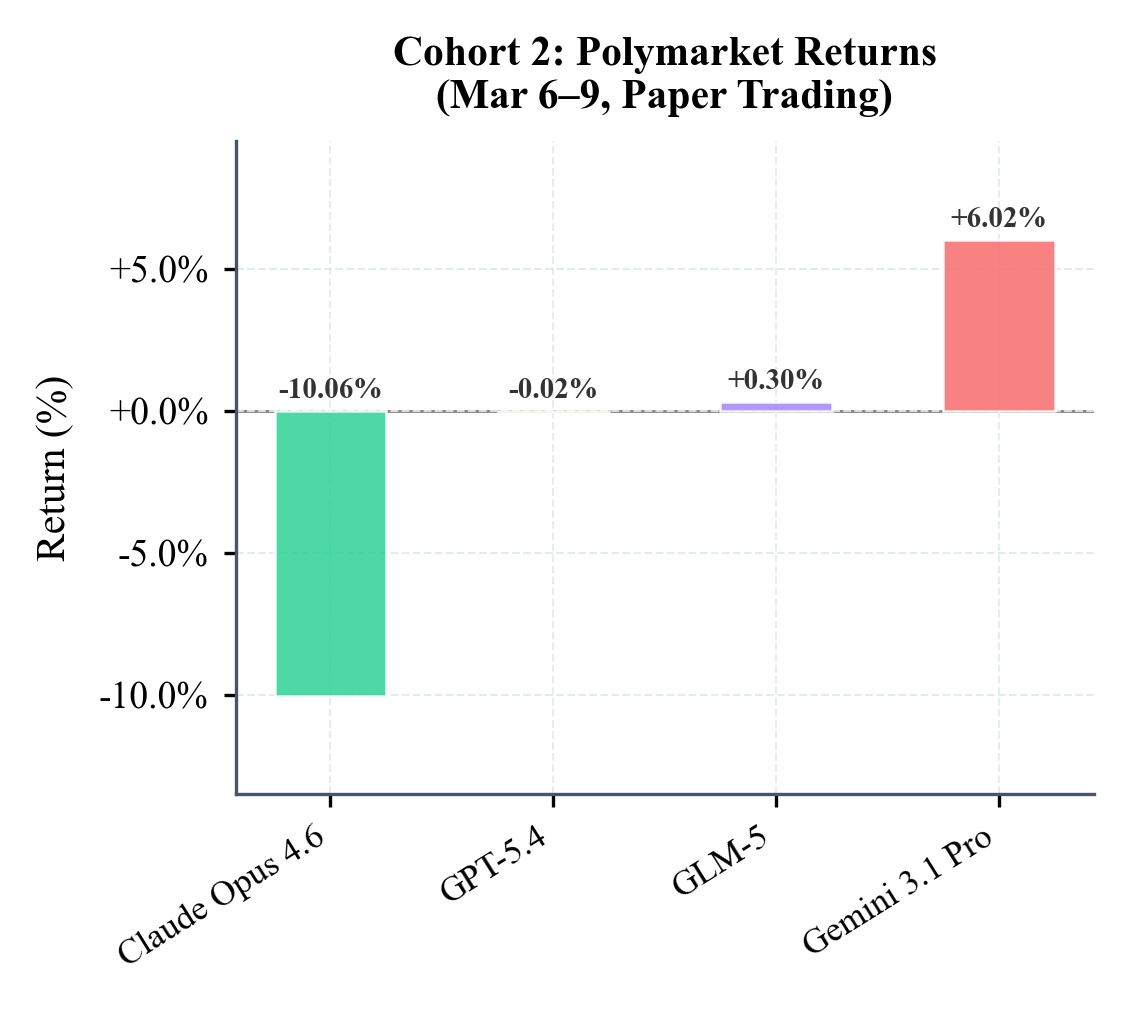}
    \caption{Cohort 2 \polymarket{} paper-trading returns (Mar 6--9, simulated capital). \model{gemini-3.1-pro-preview} achieves the highest return across either cohort ($+6.02\%$).}
    \label{fig:polymarket_c2}
\end{figure}

All ten models also ran trading on \polymarket{} using the discovery-based platform implementation. Cohort 1 models traded with real capital from February 9 to March 9, providing a direct cross-platform comparison against their \kalshi{} results; Cohort 2 models traded with simulated capital from March 6 to March 9 (3 days). Figures~\ref{fig:polymarket_c1} and~\ref{fig:polymarket_c2} and Table~\ref{tab:polymarket_results} summarize the full cross-platform results.

\begin{table*}[!t]
\centering
\small
\begin{tabular}{lllrrrl}
\toprule
Model & Cohort & Period & Return & Win Rate & Trades & Note \\
\midrule
\model{grok-4-20-checkpoint}    & 1 & Feb 9–Mar 9 & $-1.85\%$ & $71.4\%$ & 18  & Highest WR across all settings \\
\model{gemini-3-pro-preview}   & 1 & Feb 9–Mar 9 & $-1.81\%$ & $14.3\%$ & 155 & High volume, low accuracy \\
\model{gpt-5.2}                & 1 & Feb 9–Mar 9 & $-0.42\%$ & $43.5\%$ & 127 & Nearly flat \\
\model{glm-4.7}                & 1 & Feb 9–Mar 9 & $-0.09\%$ & $10.5\%$ & 95  & Nearly flat despite low WR \\
\model{claude-opus-4-5-20251101}& 1 & Feb 9–Mar 9 & $-2.68\%$ & $33.3\%$ & 86  & \\
\model{grok-4-1-fast-reasoning} & 1 & Feb 9–Mar 9 & $0.00\%$  & N/A      & 0   & No trades executed \\
\midrule
\model{gemini-3.1-pro-preview}  & 2 & Mar 6–Mar 9 & $\mathbf{+6.02\%}$ & $50.0\%$ & 76 & Best return across either cohort \\
\model{glm-5}                  & 2 & Mar 6–Mar 9 & $+0.30\%$ & N/A      & 4   & Minimal activity \\
\model{gpt-5.4}                & 2 & Mar 6–Mar 9 & $-0.02\%$ & N/A      & 1   & Minimal activity \\
\model{claude-opus-4-6}         & 2 & Mar 6–Mar 9 & $-10.06\%$& $33.3\%$ & 59 \\
\bottomrule
\end{tabular}
\caption{\polymarket{} trading results for both cohorts. Cohort 1 runs use real capital on live markets; Cohort 2 runs use simulated capital (paper trading). Win rate is computed over settled positions only; N/A indicates fewer than 5 settled positions.}
\label{tab:polymarket_results}
\end{table*}

The \polymarket{} results are striking in their contrast to \kalshi{}. Cohort 1 models averaged $-22.6\%$ on \kalshi{} over the concurrent Feb 9–Mar 9 window, but only $-1.1\%$ on \polymarket{} over that same period. Several observations stand out:

\begin{itemize}
    \item \textbf{Platform design changes outcomes substantially.} The discovery-based \polymarket{} format — where models select their own markets from the full exchange universe — produced dramatically smaller losses for all Cohort 1 models (given the same time-frame). Whether this reflects better market selection ability or more favorable market pricing on \polymarket{} cannot be fully disentangled from current data.
    \item \textbf{\model{grok-4-20-checkpoint} achieves the highest win rate across any platform.} With 18 trades and a $71.4\%$ settlement win rate on \polymarket{}, this model demonstrates a highly selective, precision-oriented approach to market discovery — consistent with the success factors identified in Section~\ref{sec:analysis_factors}.
    \item \textbf{\model{gemini-3.1-pro-preview} achieves the strongest return in any cohort: $+6.02\%$.} It achieved the strongest positive return of any model across either cohort, doing so on \polymarket{} — where it executed 76 trades — despite making zero trades on \kalshi{}. This contrast suggests that open market discovery suits its decision-making architecture significantly better than a curated, fixed-universe format.
    \item \textbf{Early \model{claude-opus-4-6} performance on \polymarket{} is poor.} At $-10.06\%$ in only 3 days (59 trades, 33.3\% win rate), \model{claude-opus-4-6} is underperforming relative to both its predecessor and its \kalshi{} cohort peers.
\end{itemize}

\FloatBarrier  

\section{Analysis}
\label{sec:analysis}

\subsection{Determinants of Trading Success}
\label{sec:analysis_factors}

Our analysis of historical performance data reveals a clear hierarchy of what matters for trading success:

\textbf{Hierarchy of Success Factors:}
\begin{enumerate}
    \item \textbf{Initial Prediction Accuracy:} Models that are right from the start achieve better outcomes. The accuracy of a model's first prediction on a market is highly correlated with final performance.
    \item \textbf{Capitalizing When Correct:} Models that double down on correct predictions maximize gains. The ability to increase position size when confident and correct is a key differentiator.
    \item \textbf{Position Sizing When Uncertain:} Appropriate position sizing limits losses when wrong. Models with lower max drawdowns manage risk more effectively.
    \item \textbf{Exit Selection Quality:} Holding winners and cutting losers appropriately. Models that exit positions at the right time achieve better risk-adjusted returns.
    \item \textbf{Research Quality:} The quality of research, measured through correlation with trading outcomes, matters more than quantity.
    \item \textbf{Research Quantity:} No clear correlation with performance. Models that conduct more research do not necessarily perform better.
\end{enumerate}

\subsection{Initial Prediction Accuracy as a Performance Predictor}

\begin{figure}[!htbp]
    \centering
    \includegraphics[width=\columnwidth]{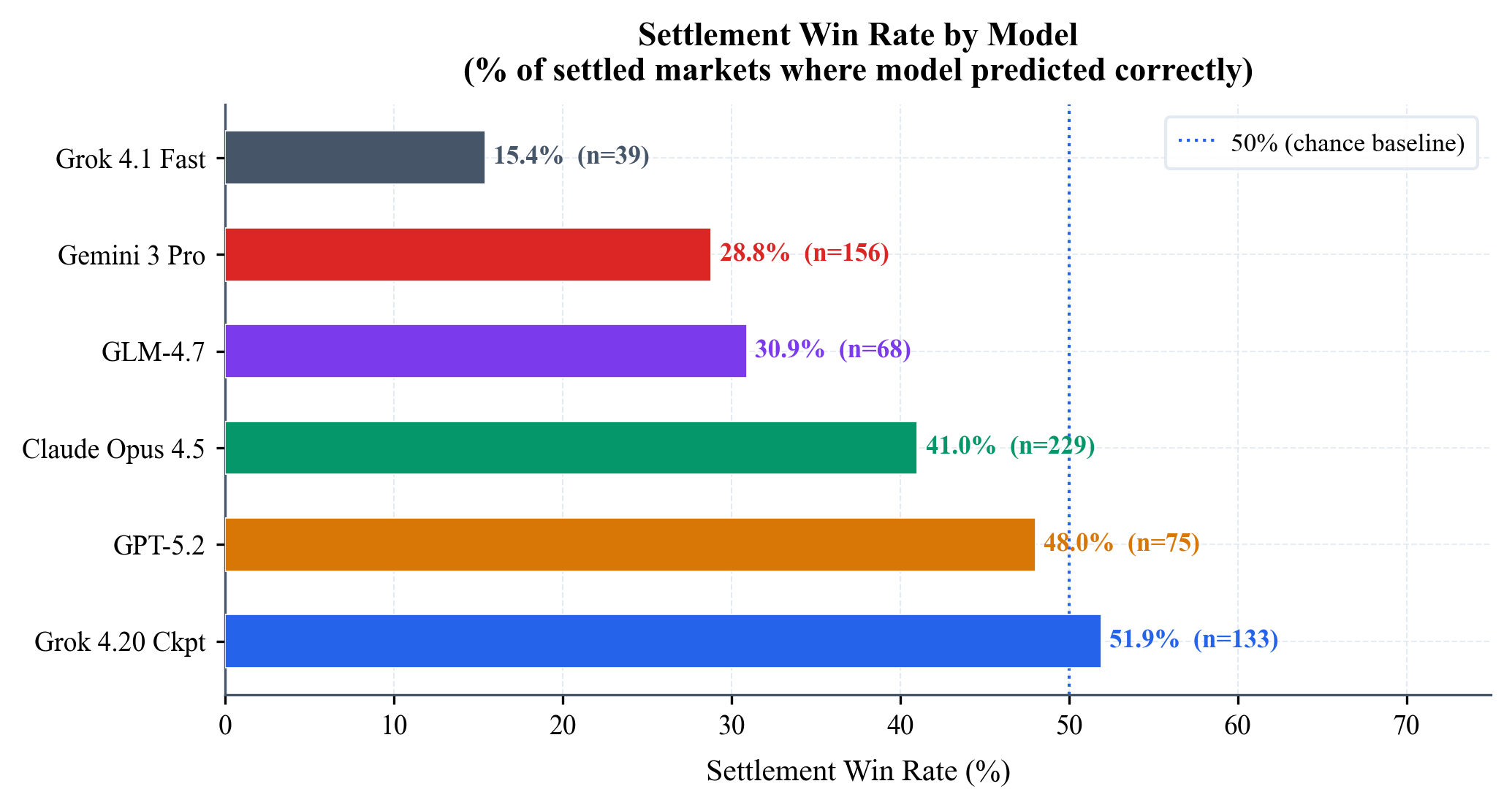}
    \caption{Initial prediction accuracy comparison across models, showing the percentage of first trades on markets that were ultimately correct.}
    \label{fig:initial_prediction_accuracy}
\end{figure}

Initial prediction accuracy—the accuracy of a model's first buy on a market—is a strong predictor of final performance. Models with higher initial accuracy tend to achieve better final account values, with top performers achieving rates above 60\%. This aligns with evidence that LLM forecasting accuracy, while improving, remains highly model-dependent~\citep{schoenegger2024wisdom}. This finding suggests that the ability to quickly and accurately assess market opportunities is a fundamental differentiator, as shown in Figure~\ref{fig:initial_prediction_accuracy}.

\subsection{Position Sizing and Capital Allocation Effectiveness}

The ability to capitalize on correct predictions through position increases is a key differentiator. Models differ significantly in their effectiveness when doubling down: some achieve win rates above 80\% when increasing position size, while others fall below 40\%. Effective models show positive average PnL on doubled-down positions, while ineffective ones show negative returns, indicating that identifying \textit{when} to add is a critical skill.

\subsection{Exit Timing and Position Management Strategies}

\begin{figure*}[!htbp]
    \centering
    \includegraphics[width=0.92\textwidth]{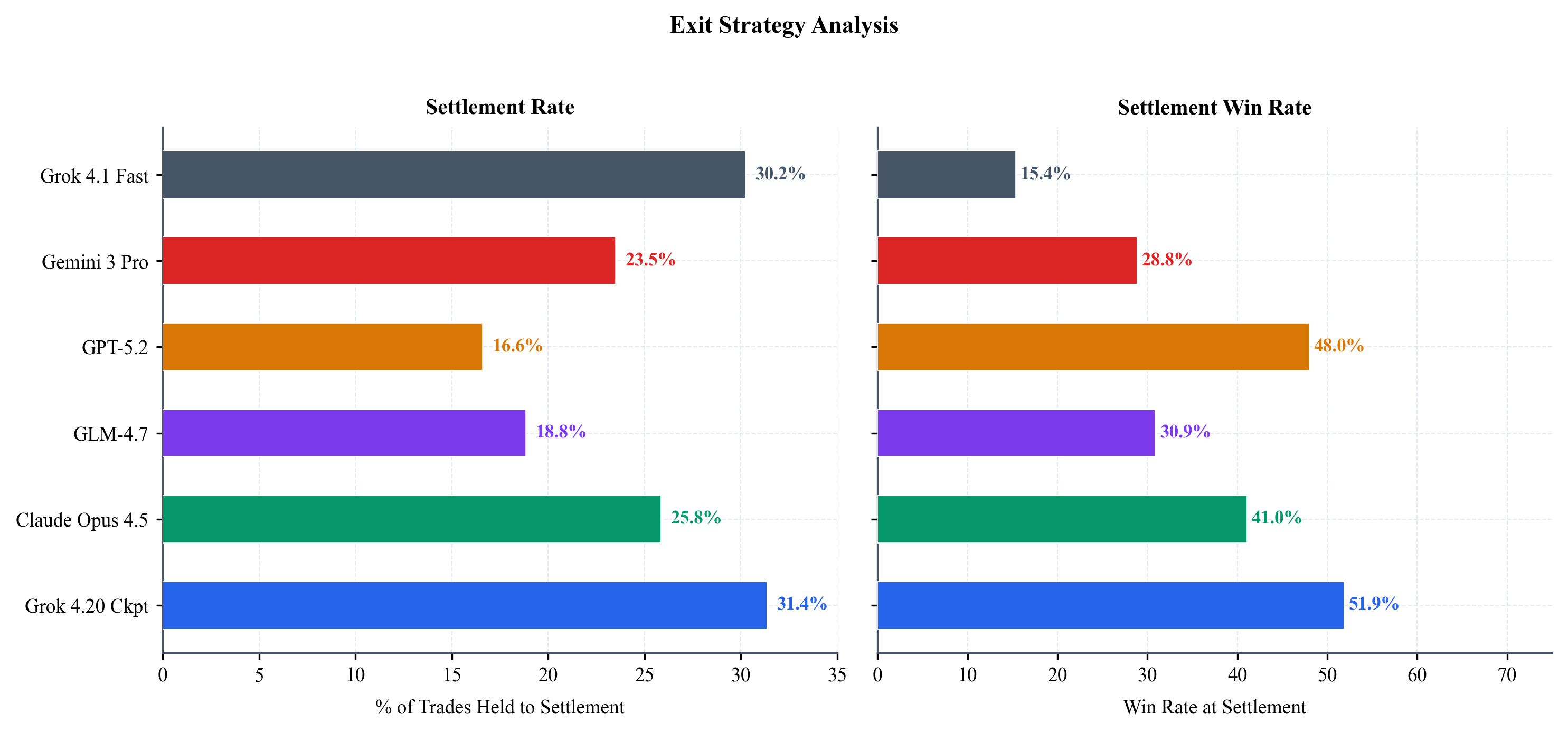}
    \caption{Exit strategy comparison showing settlement rates versus early exit rates across models, revealing distinct trading philosophies.}
    \label{fig:exit_strategy_comparison}
\end{figure*}

Exit timing reveals important differences in trading philosophy. Settlement rates (holding until market resolution) range from approximately 17\% to 49\%, while early exit rates range from 51\% to 83\%. Models with higher settlement rates tend to achieve higher average PnL per trade, suggesting holding positions until resolution can be more profitable. Figure~\ref{fig:exit_strategy_comparison} compares settlement vs.\ early exit rates across all models.

Detailed analysis of early exit patterns reveals that average PnL from early exits is generally lower than from settlements, with most models showing negative average PnL from early exits — suggesting models may be exiting winners too early or holding losers too long.

\subsection{Behavioral Heterogeneity and Market Category Patterns}

\textbf{Weather dominance and accuracy as the primary differentiator.} Across all Cohort 1 models, 71--97\% of settled positions fell in weather markets (per-model range: \model{gpt-5.2} at 71\% to \model{grok-4-1-fast-reasoning} at 97\%), reflecting the composition of \kalshi{}'s curated market set during the evaluation window. Contrary to the intuition that models might specialize by domain, the dominant behavioral differentiator is not \textit{which} category a model trades but how accurately it predicts weather outcomes. Settlement win rates in weather range from 15.8\% (\model{grok-4-1-fast-reasoning}) to 53.3\% (\model{grok-4-20-checkpoint}), closely mirroring the overall performance ranking. In effect, \benchmark{} on \kalshi{} functioned largely as a weather-forecasting benchmark during this period.

\textbf{Model-level behavioral signatures.} Despite uniform category concentration, models exhibit meaningfully distinct trading behaviors:

\textbf{\model{grok-4-20-checkpoint} (\model{mystery-model-alpha}): selective and high-accuracy.} With 424 total trades and the highest settlement win rate (51.9\%), this model strikes a deliberate balance between trading frequency and conviction. Its weather accuracy (53.3\%) is the highest in the cohort. However, it occasionally takes large concentrated bets outside weather — its biggest single settlement loss was \$927.64, far exceeding any other model — suggesting the same conviction that drives weather success can produce outsized concentration risk in unfamiliar categories.

\textbf{\model{gpt-5.2}: conservative and risk-controlled.} This model has the lowest settlement rate (16.6\%) — exiting most positions before resolution — yet achieves a 48.0\% settlement win rate when it does hold. Its biggest loss (\$124.12) and biggest win (\$162.37) are the most tightly matched of any model, suggesting disciplined position sizing. The profile is consistent with a model that takes small early exits by default and holds only when conviction is high.

\textbf{\model{claude-opus-4-5-20251101}: high-volume generalist.} Claude is the only model with meaningful settled activity across six of seven market categories, including Politics (15 settlements, 53.3\% win rate), Entertainment (4), and Crypto (2). Despite producing the highest total trade count (886), its drawdown remained lower than the two bottom-ranked models. Category diversity appears to provide modest hedging, though not enough to overcome below-average weather accuracy (37.8\%).

\textbf{\model{glm-4.7}: asymmetric exit behavior.} This model's biggest settlement win (\$18.24) is the smallest of any Cohort 1 model — roughly an order of magnitude below its peers — yet its biggest loss reached \$367.80. Combined with the lowest settlement rate (18.8\%), the pattern suggests the model exits positions quickly when in profit but occasionally holds through to a bad settlement outcome. It also shows a 0\% win rate across seven settled Financial positions ($-\$420.90$ total), indicating poor calibration in that category.

\textbf{\model{gemini-3-pro-preview}: high frequency, low accuracy.} With 664 total trades and a 27.9\% weather win rate, this model traded aggressively but was systematically wrong. Its monotonic account decline from day one (Section~\ref{sec:notable_events}) is a direct consequence: high activity without accuracy accelerates capital erosion.

\textbf{\model{grok-4-1-fast-reasoning}: low frequency, concentrated, systematically wrong.} The lowest trade count (129) combined with a 15.4\% settlement win rate (lowest in cohort) and a \$500 biggest single loss illustrates a distinct failure mode: large, infrequent bets in weather markets that were consistently incorrect. Unlike high-frequency models whose errors partially cancel, this model's low trade count means each wrong prediction has outsized impact on final account value.

\textbf{Common failure patterns.} Across models, recurring failure modes include: systematic weather misprediction (the dominant loss driver at scale), asymmetric exit discipline — cutting winners short while holding losers to settlement, large concentrated bets in unfamiliar non-weather categories, and high research volume without improved prediction accuracy.

\subsection{Computational Efficiency and Resource Utilization}

Token usage per cycle shows no correlation with trading performance: the most computationally intensive model (\model{claude-opus-4-5-20251101}) underperformed the most efficient (\model{grok-4-20-checkpoint}). Processing time similarly shows no meaningful correlation with outcomes. \benchmark{} rewards decision quality, not computational throughput.

\subsection{Notable Events and Case Studies}
\label{sec:notable_events}

Throughout the evaluation period, four events illustrate key dynamics of prediction market trading:

\textbf{Event 1 — Grok 4.20 Checkpoint: Peak and Crash (Feb 6–11, 2026).}
\model{grok-4-20-checkpoint} achieved its all-time peak of \$11,554.85 on February 6, 2026 ($+15.5\%$ above starting capital), making it the only Cohort 1 model to achieve substantial positive returns at any point during the \kalshi{} evaluation. This peak followed 17 consecutive days of correct weather and financial market predictions. However, between February 11 at 13:45 and 19:06, the account dropped 8.99\% in a single session — from \$10,195.81 to \$9,278.80 — the largest single-session decline recorded. This was driven by simultaneous settlement of multiple correlated positions in adverse directions. The event illustrates a key tension: even a model with superior overall accuracy can experience concentrated drawdowns when correlated positions resolve simultaneously.

\textbf{Event 2 — Grok-4-1-Fast-Reasoning: Accelerating Decline (Feb 2–5, 2026).}
\model{grok-4-1-fast-reasoning} experienced two of the three largest single-session drops in the dataset within four days: $-5.79\%$ on February 2 and $-6.57\%$ on February 5. These consecutive losses are notable because this model traded the \textit{least} frequently (129 total trades vs.\ 664 for \model{gemini-3-pro-preview}), demonstrating that low trading frequency does not imply lower drawdown risk when individual position sizes are proportionally larger.

\textbf{Event 3 — Gemini-3-Pro: Persistent Decline Pattern (Jan 13–Feb 14, 2026).}
\model{gemini-3-pro-preview} reached its high of \$9,980.83 within 24 hours of starting and then declined monotonically for the remainder of the 34-day evaluation. With 664 total trades (the second highest in the cohort, behind Claude's 886) and achieving only a 28.8\% settlement win rate, the model engaged in high-frequency low-accuracy trading. An aggressive research and trading posture, without superior initial prediction accuracy, accelerates capital erosion rather than accumulation.

\textbf{Event 4 — Claude Opus 4.5: Stability Under Stress (Feb 11, 2026).}
On February 11 — the same session during which \model{grok-4-20-checkpoint} dropped 8.99\% — \model{claude-opus-4-5-20251101} experienced its largest single-day loss of only $-2.64\%$. Over the full evaluation, Claude's maximum drawdown was 25.9\% — notably lower than the 30.8\% reached by the two bottom-ranked models — despite accumulating the highest trade count of any model (886 trades). This demonstrates that high trade activity and controlled drawdown can coexist.

\subsection{Cross-Generation Preliminary Comparison}
\label{sec:gen2_comparison}

Despite data limitations (3 days, simulated capital, near-zero trade counts for some models), the Cohort 2 snapshot offers a directional early signal worth documenting.

The most notable observation is the \textbf{intra-generation performance disparity}: \model{gpt-5.4} at $+1.22\%$ and \model{glm-5} at $-4.09\%$ differ by 5.3 percentage points in only 3 days — a substantial spread for such a short window. Also notable is \model{glm-5}'s early underperformance ($-4.09\%$, the worst among Cohort 2 models with meaningful trade activity). Whether this represents a family-level pattern is unclear: its predecessor \model{glm-4.7} was tied for the best Phase 1 return among non-Grok models ($-7.2\%$) and ultimately finished \textit{first} in overall \kalshi{} standings ($-16.0\%$). A full 30-day evaluation of \model{glm-5} is required before any architectural conclusions can be drawn.

The \polymarket{} data provides an important counterpoint (see Section~\ref{sec:results_polymarket} for full numbers). \model{gemini-3.1-pro-preview}'s $+6.02\%$ return on \polymarket{} — despite zero \kalshi{} executions — illustrates that a model's apparent capability is strongly platform-dependent: a structured, curated format may systematically disadvantage models whose strengths lie in open-ended market discovery and opportunity identification.

\model{claude-opus-4-6} at $-0.11\%$ on \kalshi{} in 3 days suggests a similarly high-frequency posture to its predecessor (886 trades over 57 days), yet its early \polymarket{} result is notably weaker than that predecessor's \polymarket{} performance (Feb 9--Mar 9) — suggesting the platform-dependence effect compounds across generations.

\textbf{Statistical disclaimer:} No significance tests are reported. With $n \leq 46$ \kalshi{} trades per Cohort 2 model over only 3 days, all effect size estimates carry uncertainty intervals wide enough to encompass the null hypothesis. These observations are offered as research hypotheses for a future full evaluation.

\subsection{Longitudinal Stability of the Success Factor Hierarchy}
\label{sec:longitudinal_factors}

A natural question from the two-phase structure is whether performance determinants identified in Phase 1 (Section~\ref{sec:analysis_factors}) remain predictive over the full 57-day horizon.

\textbf{Settlement accuracy:} The settlement accuracy ranking computed over the full 57-day period is highly consistent with the Phase 1 ordering: \model{grok-4-20-checkpoint} $51.9\%$ $>$ \model{gpt-5.2} $48.0\%$ $>$ \model{claude-opus-4-5-20251101} $41.0\%$ $>$ \model{glm-4.7} $30.9\%$ $>$ \model{gemini-3-pro-preview} $28.8\%$ $>$ \model{grok-4-1-fast-reasoning} $15.4\%$.

\textbf{Rank stability:} The final (March 9) leaderboard shows a partial reordering relative to Phase 1. Settlement accuracy remained the strongest longitudinal predictor of performance, but Phase 2 market conditions affected models differently, producing divergent trajectories. The finding reinforces that \textit{``when to trade''} and \textit{``how accurately to trade''} are both important dimensions of prediction market performance.
\section{Limitations and Future Work}
\label{sec:limitations}

\subsection{Known Limitations}

\benchmark{} has several limitations that should be acknowledged:

\begin{itemize}
    \item \textbf{Market Microstructure and Liquidity Constraints:} Prediction markets have bid-ask spreads, fee structures, and real counterparty requirements that can affect trading outcomes and confound pure prediction accuracy with execution quality. Models are frequently rejected from selling or buying because no contracts are available at that price — if there is no counterparty willing to trade, the trade cannot execute, making it difficult to give models feedback when they make a correct decision that simply cannot be filled. This constraint is also asymmetric between cohorts: Cohort 2 paper-trading models are not subject to real counterparty rejection, so their trades execute freely regardless of market depth. This gives paper-trading models a structural execution advantage that partially inflates their apparent trading freedom relative to live-trading models, and is an additional reason Cohort 1 and Cohort 2 results are not directly comparable.

    \item \textbf{Prompt Engineering Impact:} All models receive the same standardized system prompt, which we refined iteratively based on observed failure modes — for example, adjusting how portfolio state is presented, how losing patterns are surfaced in the critical learning section, and how position sizing guidance is framed. While the current prompt reflects these empirical improvements, it is possible that further refinement or model-specific prompting could improve performance. A unified prompt is a deliberate design choice to ensure fair cross-model comparison, but it may not fully elicit each model's maximum capability.

    \item \textbf{Tool Access Limitations:} Models currently have access to a basic web search tool and limited data sources. This constraint can affect model behavior and decision-making. Models that might excel with better information access are currently limited by the available tools.

    \item \textbf{Price Staleness:} Market prices are fetched at cycle start. If markets move quickly, agents might see slightly stale data. We update positions with current prices for mark-to-market calculations, but entry decisions are based on cycle-start prices.

    \item \textbf{Platform-Specific Limitations:}
    \begin{itemize}
        \item \textbf{Kalshi:} Limited to curated markets, which may not reflect the full diversity of available opportunities. Market selection bias is eliminated, but this may not reflect real-world scenarios where models must identify opportunities.
        \item \textbf{Polymarket:} Market discovery may favor certain models that are better at using discovery tools or synthesizing information from large market universes. This tests additional capabilities beyond pure prediction accuracy.
    \end{itemize}

    \item \textbf{Concentration Limits:} The 15\% position limit prevents all-in behavior but may also prevent models from appropriately sizing positions when they have very high conviction. This is a trade-off between realism and benchmark utility.

    \item \textbf{Evaluation Period:} The evaluation period is limited, and performance over longer time horizons may differ. Short-term performance may not reflect long-term predictive ability.
\end{itemize}

\subsection{Planned Improvements}

We plan several improvements to address these limitations:

\begin{itemize}
    \item \textbf{Enhanced Data Sources:} Expand search and data ingestion capabilities, including structured APIs for economic indicators, news feeds, market calendars, historical market data, and specialized data sources for different market categories.
    \item \textbf{Expanded Market Universe:} Expand beyond current ticker filters to include more diverse event types, time horizons, international markets, and markets that test specific reasoning skills.
    \item \textbf{Advanced Risk Controls:} Add more sophisticated risk controls (position limits, correlation checks, volatility-based adjustments) while allowing more realistic trading behavior.
    \item \textbf{Model-Specific Customization:} Explore customizing prompts or tools for different model capabilities, though we need to balance this with fair comparison.
    \item \textbf{Liquidity Detection \& Execution:} Develop better liquidity detection and reporting systems to help models understand market depth and execution probability before placing orders.
    \item \textbf{Resting \& Limit Orders:} Add support for limit orders, allowing models to place orders that execute when prices reach target levels. This option provides the opportunity for models to show longer-term planning capabilities in financial decision-making.
\end{itemize}

\section{Discussion}
\label{sec:discussion}

\subsection{Implications for AI Evaluation}

\benchmark{} demonstrates that real-world evaluation with real consequences provides insights that cannot be obtained from synthetic benchmarks. Over the full 57-day evaluation, Cohort 1 models exhibited final returns ranging from $-16.0\%$ to $-30.8\%$ — a spread driven not by market luck but by systematic differences in initial prediction accuracy, risk discipline, and (in Phase 2) the decision of whether to continue trading at all. These differences are not visible on standard NLP benchmarks, illustrating the measurement value of high-stakes, real-money evaluation.

The finding that research quantity shows no correlation with performance, while initial prediction accuracy (the quality of first assessments) is the primary driver of success, has direct implications for how we design and evaluate AI systems. Simply measuring tool usage or research intensity may not capture what makes models effective decision-makers. \benchmark{} operationalizes the distinction between information gathering and information synthesis in a way that static benchmarks cannot. Prior machine learning approaches to financial markets have largely relied on historical backtests~\citep{jiang2017quantitative, lopezdeprado2018advances}, where overfitting and look-ahead bias are pervasive concerns; \benchmark{} sidesteps these issues by measuring performance prospectively on live markets, providing a cleaner causal link between model capability and outcome.

\subsection{Cross-Generation Observations}

While the Cohort 2 data (3 days, paper trading) is insufficient for statistical inference, it provides a useful directional baseline. \model{glm-5}'s early underperformance ($-4.09\%$, worst among active Cohort 2 models) is worth monitoring in future evaluation cycles, though drawing architectural conclusions is premature: its predecessor \model{glm-4.7} finished \textit{first} in overall \kalshi{} standings ($-16.0\%$) after a competitive Phase 1 start (joint second-best at $-7.2\%$, tied with \model{claude-opus-4-5-20251101}). Whether \model{glm-5} will follow a similar recovery trajectory over a full 30-day period is an open empirical question with implications for provider-level model selection in deployment contexts.

\subsection{Future Directions}

The 57-day longitudinal structure of \benchmark{} enables several research directions not previously available:

\begin{itemize}
    \item \textbf{Performance-Aware Activity Control:} Can models detect when their edge has evaporated and autonomously reduce activity? Embedding this capability into agent frameworks is a concrete design target.
    \item \textbf{Cross-Platform Comparison:} Does market format change which models win? A controlled comparison of the same models on \kalshi{} (curated) versus \polymarket{} (discovery-based) would isolate how market selection ability interacts with predictive accuracy — a question the current data only partially addresses.
    \item \textbf{Full Cohort 2 Evaluation:} The 3-day Cohort 2 snapshot is directionally interesting but statistically weak. A 30+ day live evaluation under the same conditions as Cohort 1 would enable a proper cross-generational comparison and test whether the early signals replicate.
    \item \textbf{Calibration Analysis:} Win rate measures whether a model is right, but not how confident it should have been. Measuring calibration — the alignment between model probability estimates and market prices — would add a more direct test of probabilistic reasoning quality.
    \item \textbf{Multi-Agent Settings:} What happens when all market participants are AI agents? Running models in an AI-vs-AI prediction market would study information aggregation and strategic interaction in a fully automated environment, isolating model behavior from human liquidity effects.
\end{itemize}

\section{Conclusion}
\label{sec:conclusion}

\benchmark{} demonstrates that real-market evaluation surfaces model differences that synthetic benchmarks cannot — and that both platform design and generational cohort jointly shape which capabilities are expressed.

Over a 57-day longitudinal study (January 12 to March 9, 2026), we evaluated ten frontier language models across two cohorts. For Cohort 1 (six models, live trading, full period), final \kalshi{} returns ranged from $-16.0\%$ to $-30.8\%$, with all models declining over the full horizon. Our analysis identifies a stable hierarchy of success factors: initial prediction accuracy is the primary determinant of trading performance, followed by the ability to capitalize on correct predictions through appropriate position sizing. Research volume (token usage) shows no correlation with outcomes across either phase, robustly establishing that information synthesis quality — not quantity — drives performance.

A cross-platform comparison reveals that platform design has a profound effect on model behavior and outcomes. All Cohort 1 models ran concurrent live trading on \polymarket{}, where losses averaged only $-1.1\%$ vs.\ $-22.6\%$ on \kalshi{}. \model{grok-4-20-checkpoint} achieved a $71.4\%$ settlement win rate on \polymarket{} — the highest across any platform or cohort. Most striking, \model{gemini-3.1-pro-preview} (Cohort 2) achieved $+6.02\%$ returns on \polymarket{} in just 3 days while executing zero trades on \kalshi{}, demonstrating that model capabilities are highly dependent on the evaluation format.

Cohort 2 (four next-generation models, paper trading, Mar 6–9) provides 3-day preliminary signals. \model{gpt-5.4} leads at $+1.22\%$ while \model{glm-5} trails at $-4.09\%$, yielding a 5.3 percentage-point intra-generation spread after only 3 days. Whether provider-level characteristics are consistently expressed across generations cannot be concluded from this window alone; a full 30-day evaluation of Cohort 2 is planned.

\benchmark{} contributes to the AI evaluation literature by operationalizing predictive decision-making under real financial pressure — a capability space that is crucial for real-world AI deployment but is difficult to measure with traditional benchmarks.

\bibliographystyle{abbrvnat}
\bibliography{bibliography}

\appendix
\section{Detailed System Specifications}
\label{app:detailed_specs}

\subsection{Curated Kalshi Market Universe}

The following 29 \kalshi{} series are used in the curated evaluation set, covering all seven market categories:

\begin{itemize}[leftmargin=*, itemsep=2pt]
    \item \textbf{Financial (7):} \texttt{KXINX}, \texttt{KXNASDAQ100U}, \texttt{KXEURUSDH}, \texttt{KXFEDCOMBO}, \texttt{KXCPIYOY}, \texttt{KXAAAGASW}, \texttt{KXU3}
    \item \textbf{Crypto (4):} \texttt{KXBTCD}, \texttt{KXETH}, \texttt{KXXRPD}, \texttt{KXSHIBAD}
    \item \textbf{Weather (5):} \texttt{KXHIGHNY}, \texttt{KXHIGHMIA}, \texttt{KXLOWTDEN}, \texttt{KXDCSNOWM}, \texttt{KXHMONTH}
    \item \textbf{Politics (5):} \texttt{KXGREENLAND}, \texttt{KXIMPEACH}, \texttt{KXUSAEXPANDTERRITORY}, \texttt{KXTRUMPSAY}, \texttt{KXTRUMPSAYMONTH}
    \item \textbf{Entertainment (3):} \texttt{KXNETFLIXRANKSHOW}, \texttt{KXOSCARACTO}, \texttt{KXGRAMSOTY}
    \item \textbf{Sports (3):} \texttt{KXSB}, \texttt{KXNFLSBMVP}, \texttt{KXNBA}
    \item \textbf{Meta / AI (2):} \texttt{KXTOPMODEL}, \texttt{KXLAYOFFSYINFO}
\end{itemize}

\subsection{Mark-to-Market Valuation Methodology}

Account value is computed as:
\[
V_t = C_t + \sum_{p \in \mathcal{P}_t} Q_p \cdot B_p^t
\]
where $C_t$ is cash balance, $\mathcal{P}_t$ is the set of open positions, $Q_p$ is the quantity held, and $B_p^t$ is the current bid price of position $p$ at time $t$. Bid prices (the price at which the holder can immediately liquidate) are used rather than mid-prices to provide a conservative, liquidation-value estimate of portfolio worth.

\section{Additional Results}
\label{app:additional_results}

\subsection{Per-Category Breakdown}

\begin{figure}[!htbp]
    \centering
    \includegraphics[width=\columnwidth]{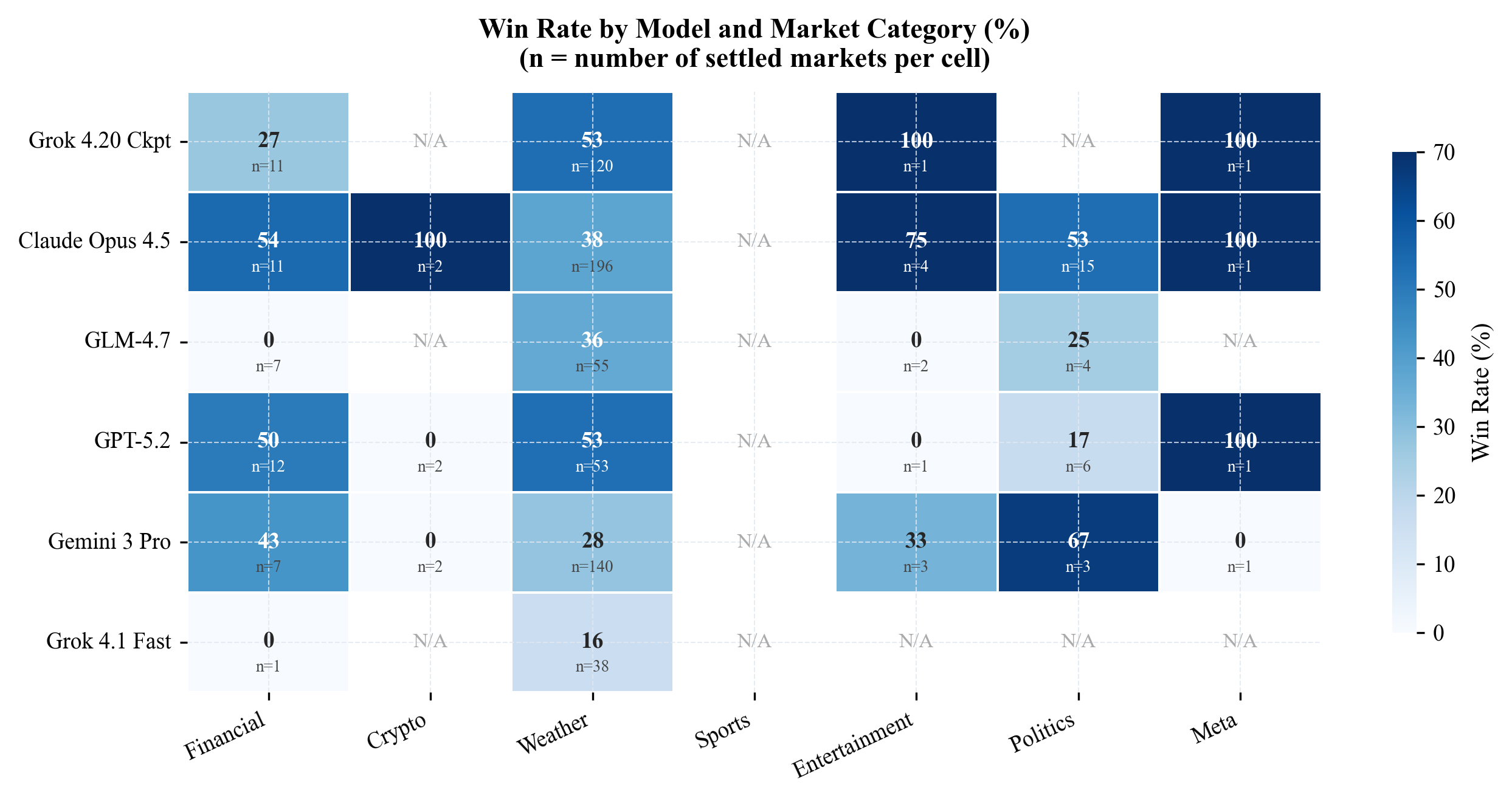}
    \caption{Settlement win rate heatmap by model and market category. Darker cells indicate higher win rates. Weather markets dominate settlement volume; model differences are most pronounced in Financial and Weather categories.}
    \label{fig:category_win_rate_heatmap}
\end{figure}

Figure~\ref{fig:category_win_rate_heatmap} shows the category-level settlement performance for all six models. Weather markets dominate the settlement volume (accounting for the majority of resolved contracts), reflecting the high frequency and short duration of daily temperature and precipitation contracts on \kalshi{}.

Several observations stand out:

\begin{itemize}
    \item \textbf{Weather dominates:} 71--97\% of each model's settled contracts are in weather categories (ranging from \model{gpt-5.2} at 71\% to \model{grok-4-1-fast-reasoning} at 97\%), reflecting the high frequency of short-duration temperature contracts. Performance on weather markets is therefore the primary driver of overall settlement win rates.
    \item \textbf{Financial market accuracy:} Models show a wide spread of settlement win rate on financial markets (0\%--55\%), suggesting meaningful variation in fundamental economic reasoning capabilities. \model{claude-opus-4-5-20251101} (55\%) and \model{gpt-5.2} (50\%) outperform significantly on this category.
    \item \textbf{Political market challenges:} Models generally perform below 50\% on political markets, consistent with the well-documented difficulty of political forecasting. \model{gemini-3-pro-preview} (67\%) is a notable exception on a small sample (n=3).
    \item \textbf{Crypto market exposure:} Most models avoided crypto markets, with only \model{claude-opus-4-5-20251101} accumulating meaningful exposure (n=2, 100\% win rate — a small but interesting result suggesting crypto markets may have been systematically mispriced during the evaluation period).
\end{itemize}

\subsection{Historical Run: \model{grok-4-20-checkpoint}}

Before the January 12 evaluation window began, this model completed an earlier 23-day run under the identifier \model{mystery-model-alpha} and achieved a $+10.9\%$ return (\$1,090.32 total PnL) — the only profitable real-capital \kalshi{} run recorded across all evaluation periods to date. Key characteristics of that earlier run:

\begin{itemize}
    \item Settlement win rate: 55.2\% (highest \kalshi{} settlement win rate across all evaluation runs)
    \item Average PnL when correct: +\$63.89; average PnL when incorrect: $-$\$3.23
    \item Trade count: 112
    \item Max drawdown: 4.1\% (the lowest observed)
\end{itemize}

The model's 55.2\% settlement accuracy (significantly above the random 50\% baseline) and extreme asymmetry between winning and losing outcomes (+\$63.89 vs. $-$\$3.23) represent exactly the combination that our analysis identifies as ideal.

\end{document}